\documentclass[preprint,12pt]{elsarticle}

\usepackage{lineno}
\usepackage{xcolor}
\usepackage{graphicx}
\usepackage{amssymb}
\usepackage{footnote}
\usepackage{amsmath}
\usepackage[ruled,vlined]{algorithm2e}
\usepackage{mathrsfs}

\usepackage{xcolor}
\definecolor{Blue}{rgb}{0,0,0.6}
\usepackage{cancel}
\usepackage{comment}
\usepackage[colorlinks = true, allcolors = Blue]{hyperref} 
\usepackage{booktabs}
\usepackage{dblfloatfix}
\usepackage{color, colortbl}
\usepackage{mathtools}
\usepackage{hhline}
\usepackage{amsfonts}
\usepackage{array}
\usepackage{soul}
\usepackage{subcaption}
\usepackage{amsthm}
\newcolumntype{L}[1]{>{\raggedright\let\newline\\\arraybackslash\hspace{0pt}}m{#1}}
\newcolumntype{C}[1]{>{\centering\let\newline\\\arraybackslash\hspace{0pt}}m{#1}}
\newcolumntype{R}[1]{>{\raggedleft\let\newline\\\arraybackslash\hspace{0pt}}m{#1}}

\makeatletter
\def\ps@pprintTitle{%
  \let\@oddhead\@empty
  \let\@evenhead\@empty
  \def\@oddfoot{\reset@font\hfil\thepage\hfil}
  \let\@evenfoot\@oddfoot
}
\makeatother

\makeatletter
\def\maketag@@@#1{\hbox{\m@th\normalfont\normalsize#1}}
\makeatother

\SetKwInput{KwInput}{Input}                
\SetKwInput{KwOutput}{Output}           

\begin{document}

\begin{frontmatter}

\title{Reliable Neural Networks for Regression Uncertainty Estimation\tnoteref{mytitlenote}}



\author[mymainaddress]{Tony Tohme\corref{mycorrespondingauthor}}
\cortext[mycorrespondingauthor]{Corresponding author}
\ead{tohme@mit.edu}

\author[mysecondaryaddress]{Kevin Vanslette}
\ead{kevin.vanslette@raytheon.com}

\author[mymainaddress]{Kamal Youcef-Toumi}
\ead{youcef@mit.edu}

\address[mymainaddress]{Department of Mechanical Engineering, Massachusetts Institute of Technology, Cambridge, MA 02139} 
\address[mysecondaryaddress]{Raytheon BBN Technologies, Cambridge, Massachusetts 02138}

\begin{abstract}
While deep neural networks are highly performant and successful in a wide range of real-world problems, estimating their predictive uncertainty remains a challenging task. To address this challenge, we propose and implement a loss function for regression uncertainty estimation based on the Bayesian Validation Metric (BVM) framework while using ensemble learning. The proposed loss reproduces maximum likelihood estimation in the limiting case. A series of experiments on in-distribution data show that the proposed method is competitive with existing state-of-the-art methods. Experiments on out-of-distribution data show that the proposed method is robust to statistical change and exhibits superior predictive capability.


\end{abstract}


\end{frontmatter}
\noindent\vspace{-27pt}\\
\section{Introduction}
\label{intro}
The proven utility of accurate data analysis has caused machine learning (ML) and deep neural networks (NNs) to emerge as crucially important tools in academia, industry, and society \cite{lecun2015}. NNs have many documented successes in a wide variety of critical domains such as natural language processing \cite{collobert2008, mikolov2013, sutskever2014}, computer vision \cite{krizhevsky2012}, and speech recognition \cite{hinton2012, hannun2014}. Given their impressive performance and high accuracies, researchers have been interested in reliably deploying deep NNs in safety-critical, real-world applications. The main aspect that differentiates ML methods from traditional statistical modeling techniques is their ability to provide tractable analysis on large and informationally dense datasets. 
As the amount of data being produced each year continues to accelerate, ML-based techniques are expected to dominate the future of data analysis.\\

Unless NN models are trained in a way to make predictions that indicate uncertainty when they are not confident, these models can make overly confident, yet incorrect, predictions \cite{guo2017calibration}. While these models can guarantee a level of accuracy for data that is statistically similar to the data they trained on, they have no guarantee to make accurate predictions on statistically different (known as out-of-distribution \cite{hendrycks2016baseline}) data. For instance, after training a vanilla NN to classify the hand written digits in the MNIST dataset, one observes (far more often than not) that feeding the NN a uniformly randomly generated image results in a prediction probability one for the predicted digit. This overly certain and wrong prediction is in stark contrast with what the modeler would desire, e.g. a uniform distribution that indicates uncertainty \cite{sensoy2018evidential}.\\

The field of probabilistic machine learning seeks to avoid overly confident, yet incorrect, predictions by quantifying and estimating the predictive uncertainty of NN models \cite{krzywinski2013, ghahramani2015}. Given that these models are being integrated into real decision systems (e.g. self-driving vehicles, infrastructure control, medical diagnosis, etc.), a decision system should incorporate the uncertainty of a prediction to avoid ill-informed choices or reactions that could potentially lead to heavy or undesired losses \cite{amodei2016concrete}.\\

Deep NNs and predictive uncertainty estimation have recently been frequently applied to several applications in reliability engineering and system safety. For instance, Bayesian NNs are used in \cite{moradi2022integration} for operation condition and risk monitoring of complex engineering systems. In \cite{zou2021resilience}, a novel deep-ensemble-assisted active learning approach is presented and developed for resilience-based recovery scheduling of transportation networks in a mixed traffic environment (i.e. connected-autonomous and human-driven vehicles). In \cite{liu2021artificial}, the authors propose an NN supported stochastic process for degradation modeling and prediction. In \cite{seo2022deep}, an NN-based optimization framework is proposed for safety evacuation route during toxic gas leak incidents. Other notable works employ deep NNs and uncertainty quantification in structural reliability analysis \cite{afshari2022machine}, assets reliability \cite{izquierdo2019dynamic}, remaining useful lifetime predictions of multi-component systems \cite{nguyen2022probabilistic}, reliability analysis of detecting false alarms in wind turbines \cite{marugan2019reliability}, waterway risk analysis \cite{zhang2020towards}, and fault diagnosis \cite{zhou2022towards}. The aforementioned works demonstrate the importance of probabilistic machine learning in reliability engineering and safety-critical, real-world applications.\\

A comparative review of the progress made regarding predictive uncertainty estimation in NNs may be found in \cite{snoek2019}. Most of the early proposed approaches are Bayesian in nature \cite{bernardo2009bayesian}. These methods assign prior distributions to NNs' parameters and the training process updates these distributions to the ``learned" posterior distributions. The residual uncertainty in the posterior distribution of the parameters allow the network to estimate predictive uncertainty. Several methods were suggested for learning Bayesian NNs including Laplace approximation \cite{mackay1992}, Hamiltonian methods \cite{springenberg2016bayesian}, Markov Chain Monte Carlo (MCMC) methods \cite{neal1996}, expectation propagation \cite{jylanki2014expectation}, and variational inference \cite{graves2011practical}. Implementing Bayesian NNs is generally difficult and training them is computationally expensive. Recent state-of-the-art methods for predictive uncertainty estimation include probabilistic backpropagation (PBP) \cite{hernandez2015}, Monte Carlo dropout (MC-dropout) \cite{gal2016}, Deep Ensembles \cite{lakshminarayanan2017}, Bayesian Deep Ensembles \cite{he2020bayesian}, and evidential learning (for classification \cite{sensoy2018evidential} and regression \cite{amini2020deep}). While these methods mainly operate during training, other works \cite{guo2017calibration, levi2019evaluating, laves2021recalibration, rahimi2020post} proposed calibrating NNs' predictive uncertainty in a post-hoc step after training, which is critical for real-world deployment.\\



Average generalization error captures the expected ability of a model to generalize to new in-distribution data due to the i.i.d. nature of train-test data split. Among several of the error functions that can be used, log of the predictive probabilities takes the predictive uncertainty into account while assessing the error. Thus, the log of the predictive probabilities is typically used for assessing the quality of predictive methods that quantify uncertainty in regression problems \cite{nix1994, lakshminarayanan2017}. In particular, the authors in \cite{nix1994} proposed a maximum likelihood formulation to train NNs with two outputs instead of one (to learn the mean and variance) by minimizing the negative log-likelihood (NLL) loss function. In \cite{lakshminarayanan2017}, it was shown that training an ensemble of such NNs further improves predictive uncertainty estimation. As accurate measure of uncertainty is crucial for safety-critical applications, many methods have been proposed for calibration of NNs \cite{guo2017calibration, levi2019evaluating, laves2021recalibration, rahimi2020post}. Specifically, the authors in \cite{levi2019evaluating, laves2021recalibration} proposed scaling the variance where the scaling parameter is learned during the process (to achieve optimal calibration). It is worth noting that all the methods that use the maximum likelihood formulation (and hence minimize the NLL loss) typically assume the likelihood to be the Gaussian distribution. Recently, it was shown that using a heavy-tailed distribution such as the Laplace distribution improves the robustness to outliers \cite{nair2022maximum}, which is also critical for real-world deployment.\\

The Bayesian Validation Metric (BVM) was shown to be a general model validation and testing tool \cite{vanslette2020}. It measures the probability of agreement between model outputs and observed data according to a user-specified Boolean function defining model-data agreement conditions. The BVM has the ability to represent all the existing validation metrics as special cases \cite{liu2011toward}, and was shown to generalize Bayesian regression and model testing \cite{tohme2020}. As we will show later in this paper, the BVM framework can also generalize the maximum likelihood formulation for predictive uncertainty estimation in NNs. We will show the flexibility and versatility of the BVM framework in generating a novel loss function (under some assumption on the Boolean agreement condition) that can be used for predictive uncertainty estimation in NNs. In other words, NNs can be trained by maximizing the BVM probability of agreement or equivalently, minimizing the negative log probability of agreement. Although we will focus on the Gaussian likelihood in our analyses, some other likelihoods can be adopted in a similar fashion (e.g. Laplace distribution as in \cite{nair2022maximum}). Similarly, while we will focus on one particular loss function (i.e. under one particular assumption on the Boolean agreement condition), other loss functions can be derived from the BVM probability of agreement formulation by adopting different agreement conditions.\\ 
 


\emph{Contributions:} 
We present a new approach to quantify predictive uncertainty in NNs for regression tasks based on the Bayesian Validation Metric (BVM) framework proposed in \cite{vanslette2020}. Using this framework, we propose a new loss function and use it to train an ensemble of NNs (inspired by Deep Ensembles \cite{lakshminarayanan2017}). The proposed loss function reproduces maximum likelihood estimation in the limiting case. Our method is very simple to implement and only requires minor changes to the standard NN training procedure. We assess our method both qualitatively and quantitatively through a series of experiments on toy and real-world datasets, and show that our approach provides well-calibrated uncertainty estimates and is competitive with the existing state-of-the-art methods (when tested on in-distribution data). We introduce and utilize the concept of ``outlier train-test splitting'' to evaluate a method's predictive ability on out-of-distribution examples whenever their presence in a dataset is not guaranteed. We show that our method has superior predictive power compared to Deep Ensembles \cite{lakshminarayanan2017} when tested on out-of-distribution (outlier) samples. As the statistics of training datasets often differ from the statistics of the environment of deployed systems, our method can be used to improve safety and decision-making in the deployed environment by better estimating \mbox{out-of-distribution uncertainty.} 





\enlargethispage{\baselineskip}
\section{The Bayesian Validation Metric for predictive uncertainty estimation}

\subsection{Notation and problem setup}
Consider the following supervised regression task. We are given a dataset $\mathcal{D} = \{\mathbf{x}_n, t_n\}_{n=1}^N$, consisting of $N$ i.i.d. paired examples, where $\mathbf{x}_n \in \mathbb{R}^d$ represents the $n^{\text{th}}$ $d$-dimensional feature vector and $t_n \in \mathbb{R}$ denotes the corresponding continuous target variable (or label). We aim to learn the probabilistic distribution $\rho(t|\mathbf{x})$ \mbox{over the targets $t$ for given inputs $\mathbf{x}$ using NNs.}

\subsection{Maximum likelihood estimation}
In regression tasks, it is common practice to train a NN with a single output node (corresponding to the predicted mean), say $\mu(\mathbf{x})$, such that the network parameters (or weights) are optimized by minimizing the mean squared error (MSE) cost (or loss) function, \mbox{expressed as}
\begin{align}
\mathcal{C}_{\text{MSE}}=\frac{1}{N}\sum_{n=1}^N\big(t_n-\mu(\mathbf{x}_n)\big)^2.\label{mse}
\end{align}
Note that the network output $\mu(\mathbf{x})$ can be thought of as an estimate of the true mean of the noisy target distribution for a given input feature \cite{nix1994}. However, this does not take into account the uncertainty or noise in the data.\\

To capture predictive uncertainty, an alternative approach based on maximum likelihood was proposed in \cite{nix1994}, and it consists of adding another node to the output layer of the neural network, $\sigma^2(\mathbf{x})$, that estimates the true variance of the target distribution. In other words, we train a network with two nodes in its output layer: $\big(\mu(\mathbf{x}), \sigma^2(\mathbf{x})\big)$. By assuming the target values $t_n$ to be drawn from a Gaussian distribution with the predicted mean $\mu(\mathbf{x}_n)\equiv\mu_n$ and variance $\sigma^2(\mathbf{x}_n)\equiv\sigma^2_n$, we can express the likelihood $\rho(t_n|\mathbf{x}_n)$ of observing the target value $t_n$ given the input vector $\mathbf{x}_n$ as follows:
\begin{align}
\rho(t_n|\mathbf{x}_n) &\equiv \rho\big(t_n\big|\mu_n, \sigma^2_n\big)\notag\\[\smallskipamount]
&=\displaystyle \frac{1}{\sqrt{2\pi\sigma^2_n}}\,\text{exp}\,\Bigg\{-\frac{\big(t_n-\mu_n\big)^2}{2\sigma^2_n}\Bigg\}.\label{maxlikelihood}
\end{align}
The aim is to train a network that infers $\big(\mu(\mathbf{x}), \sigma^2(\mathbf{x})\big)$ by maximizing the likelihood function in (\ref{maxlikelihood}). This is equivalent to minimizing its negative log-likelihood, expressed as
\begin{align}
-&\log{\rho\big(t_n\big|\mu_n, \sigma^2_n\big)} = \frac{1}{2}\log 2\pi\sigma^2_n + \frac{\big(t_n-\mu_n\big)^2}{2\sigma^2_n}.\label{nllmle}
\end{align} 
Hence the overall negative log-likelihood (NLL) cost function is given by
\begin{align}
\mathcal{C}_{\text{NLL}} = \frac{1}{N}\sum_{n=1}^N \Bigg(\frac{1}{2}\log 2\pi\sigma^2_n + \frac{\big(t_n-\mu_n\big)^2}{2\sigma^2_n}\Bigg).\label{nll}
\end{align}
Note that $\sigma^2(\mathbf{x}) > 0$; we impose this positivity constraint on the variance by using the \emph{sigmoid} function (instead of \emph{softplus} \cite{lakshminarayanan2017} or log variance \cite{kendall2017uncertainties}) as our data will be standardized \cite{he2020bayesian}.


\subsection{The Bayesian Validation Metric}
\label{BVMlossfctn}
The Bayesian Validation Metric (BVM) is a general model validation and testing tool that was shown to generalize Bayesian model testing and regression \cite{vanslette2020, tohme2020}. The BVM measures the probability of agreement $A$ between the model $M$ and the data $D$ given the Boolean agreement function $B$, denoted as $0\leq p(A|M,D,B)\leq1$. The probability of agreement is
\begin{align}
p(A|M,D,B) =  \int_{\hat{y},y}\rho(\hat{y}|M)\cdot\Theta\big(B(\hat{y},y)\big)\cdot\rho(y)\,d\hat{y}\,dy,\label{bvmpa}
\end{align}
where $\hat{y}$ and $y$ correspond to the model output and observed data respectively, $\rho(\hat{y}|M)$ is the probability density function (pdf) representing the model predictive uncertainty, $\rho(y)$ is the data uncertainty pdf, and $\Theta\big(B(\hat{y},y)\big)$ is the indicator function of the Boolean $B$ that defines the meaning of \textit{model-data agreement}. The indicator function behaves as a probabilistic kernel between the data and \mbox{model prediction pdfs.}

\subsection{The BVM reproduces the NLL loss as a special case}
We show that the BVM is capable of replicating the maximum likelihood NN framework by representing the NLL cost function $\mathcal{C}_{\text{NLL}}$ described in (\ref{nll}) as a special case. In terms of the BVM framework, the maximum likelihood formulation is achieved by modeling the predictions using a Gaussian likelihood given by
\begin{align}
\rho\big(\hat{y}_n|M(\mathbf{x}_n)\big)&\equiv\rho\big(\hat{y}_n\big|\mu_n, \sigma^2_n\big)\notag\\[\smallskipamount] 
&=\displaystyle \frac{1}{\sqrt{2\pi\sigma^2_n}}\,\text{exp}\,\Bigg\{-\frac{\big(\hat{y}_n-\mu_n\big)^2}{2\sigma^2_n}\Bigg\},\label{gauss}
\end{align}
and by assuming the target variables to be deterministic, i.e. $\rho(y_n) = \delta(y_n-t_n)$, where $\delta(\cdot)$ is the Dirac delta function. In addition, the Boolean agreement function is defined such that the model and the data are required to ``agree exactly'' (as is the case with Bayesian model testing \cite{vanslette2020, tohme2020thesis}), and is given by $\Theta\big(B(\hat{y}_n,y_n)\big)\equiv\delta(\hat{y}_n-y_n)$, when $p(A|M,D,B)\rightarrow \rho(A|M,D,B)$ is a probability density. For a particular input feature vector $\mathbf{x}_n$, the probability density of agreement between the model and data is equal to
\begin{align}
\rho(A|M,D,B,\mathbf{x}_n) &= \int_{\hat{y}_n,y_n}\hspace{-10pt}\rho\big(\hat{y}_n|M(\mathbf{x}_n)\big)\cdot\Theta\big(B(\hat{y}_n,y_n)\big)\cdot\rho(y_n)\,d\hat{y}_n\,dy_n\notag\\[\smallskipamount]
&=\int_{\hat{y}_n,y_n}\hspace{-10pt}\rho\big(\hat{y}_n\big|\mu_n, \sigma^2_n\big)\cdot\delta(\hat{y}_n-y_n)\cdot\delta(y_n-t_n)\,d\hat{y}_n\,dy_n\notag\\[\smallskipamount]
&=\int_{\hat{y}_n}\rho\big(\hat{y}_n\big|\mu_n, \sigma^2_n\big)\cdot\delta(\hat{y}_n-t_n)\,d\hat{y}_n\notag\\[\smallskipamount]
&=\rho\big(t_n\big|\mu_n, \sigma^2_n\big)\notag\\[\smallskipamount]
&=\displaystyle \frac{1}{\sqrt{2\pi\sigma^2_n}}\,\text{exp}\,\Bigg\{-\frac{\big(t_n-\mu_n\big)^2}{2\sigma^2_n}\Bigg\}.
\end{align}
Maximizing the BVM probability density of agreement is equivalent to minimizing its negative log-likelihood,
\begin{align}
-\log{\rho(A|M,D,B,\mathbf{x}_n)} &= -\log{\rho\big(t_n\big|\mu_n, \sigma^2_n\big)}\label{nllbvm}
\end{align} 
which is Equation (\ref{nllmle}). 
Therefore, the negative log-likelihood BVM cost function over the set of all input feature vectors $\mathbf{x} = \{\mathbf{x}_1, \hdots, \mathbf{x}_N\}$ is given by
\begin{align}
\mathcal{C}_{\text{BVM}} &= -\frac{1}{N}\log{\rho(A|M,D,B,\mathbf{x})}\notag\\[\smallskipamount] 
&= -\frac{1}{N}\log{\textstyle\prod_{n=1}^N \rho(A|M,D,B,\mathbf{x}_n)}\notag\\[\smallskipamount] 
&=  \frac{1}{N}\sum_{n=1}^N -\log{\rho\big(t_n\big|\mu_n, \sigma^2_n\big)}\notag\\[\smallskipamount]
&= \mathcal{C}_{\text{NLL}},
\end{align}
which is Equation (\ref{nll}). 
Thus, with the assumptions put on the data, model, and agreement definition, the BVM method can reproduce the maximum likelihood method as a special case. That is, minimizing the BVM negative log-probability density of agreement is mathematically equivalent to minimizing the NLL loss, which was essentially \mbox{used in Deep Ensembles \cite{lakshminarayanan2017}.}

\subsection{The $\boldsymbol{\epsilon}$-BVM loss: a relaxed version of the \mbox{NLL loss}}
\label{theepseBVMlossfctn}
We now consider the $\epsilon$-Boolean agreement function $B(\hat{y}_n, y_n, \epsilon)$ being true iff $|\hat{y}_n-y_n|\leq \epsilon$. In the limit $\epsilon\to 0$, this Boolean function requires the model output and data to ``agree exactly'', which leads to the maximum likelihood NN limit of the BVM discussed above. Again, assuming the model predictive uncertainty to be Gaussian, the target variables to be deterministic and the agreement function to be $B(\hat{y}_n, y_n, \epsilon)$, the $\epsilon$-BVM probability of agreement for a given input feature vector $\mathbf{x}_n$ can be expressed as 
\begin{align}
p\big(A\big|M,D,B(\epsilon),\mathbf{x}_n\big) &=\int_{\hat{y}_n,y_n}\hspace{-10pt}\rho\big(\hat{y}_n\big|\mu_n, \sigma^2_n\big)\cdot\Theta\big(|\hat{y}_n-y_n|\leq \epsilon\big)\cdot\delta(y_n-t_n)\,d\hat{y}_n\,dy_n\notag\\[\smallskipamount]
&=\int_{\hat{y}_n}\rho\big(\hat{y}_n\big|\mu_n, \sigma^2_n\big)\cdot\Theta\big(|\hat{y}_n-t_n|\leq \epsilon\big)\,d\hat{y}_n\notag\\[\smallskipamount]
&=\int_{t_n-\epsilon}^{t_n+\epsilon}\rho\big(\hat{y}_n\big|\mu_n, \sigma^2_n\big)\,d\hat{y}_n\notag\\[\smallskipamount]
&= \Phi\bigg(\frac{t_n+\epsilon-\mu_n}{\sigma_n}\bigg) - \Phi\bigg(\frac{t_n-\epsilon-\mu_n}{\sigma_n}\bigg),\label{deltacdf}
\end{align}
where $\Phi(\cdot)$ is the cumulative distribution function (cdf) of the standard normal distribution. Thus, this $\epsilon$-BVM probability of agreement becomes the difference in likelihood cdfs around the mean. Taking its (overall) negative log gives
\begin{align}
&\mathcal{C}_{\text{BVM}}\big(B(\epsilon)\big) = \frac{1}{N}\sum_{n=1}^N -\log{p\big(A|M,D,B(\epsilon),\mathbf{x}_n\big)}\notag\\[\smallskipamount]
&\hspace{-4pt} = \frac{1}{N}\sum_{n=1}^N -\log\Bigg[\Phi\bigg(\frac{t_n+\epsilon-\mu_n}{\sigma_n}\bigg) - \Phi\bigg(\frac{t_n-\epsilon-\mu_n}{\sigma_n}\bigg)\Bigg].\label{bvmcdf}
\end{align}
Having this looser definition of model-data agreement effectively coarse-grains the in-distribution data and prevents overfitting. While Section \ref{realworlddatasets} shows that this coarse-graining increases the bias of the in-distribution test results, Section \ref{oodgen} shows that our method better generalizes to out-of-distribution sample predictions.

%

\subsection{Implementation and ensemble learning}
Implementing our proposed method is straightforward and requires little modifications to typical NNs. 
We simply train a NN using the BVM loss function. Since our aim is to estimate and quantify the predictive uncertainty, our NN will have two nodes in its output layer, corresponding to the predicted mean $\mu(\mathbf{x})$ and variance $\sigma(\mathbf{x})^2$, as we mentioned earlier. More details about our NN architecture will be discussed in the next section.\\

Training an ensemble of NNs independently and statistically integrating their results was shown to improve predictive performance \cite{lakshminarayanan2017}. This class of ensemble methods is known as a randomization-based approach (such as random forests \cite{breiman2001}) in contrast to a boosting-based approach where NNs are trained sequentially. Due to the randomized and independent training procedure, the local minima the NNs settle into vary across the ensemble. This causes the ensemble to ``agree" where there is training data and ``disagree" elsewhere, which increases the variance of the statistically integrated predictive distribution.\\



We follow \cite{lakshminarayanan2017} and adopt their ensemble learning procedure by training an ensemble of $K$ NNs, but instead we utilize the BVM loss rather than the NLL loss (recall the NLL is a special case of the BVM). We let each network $k$ parametrize a distribution over the outputs, i.e. $\rho_{\theta_k}(t|\mathbf{x},\theta_k)$ where $\theta_k$ represents the vector of weights of network $k$. In addition, we assume the ensemble to be a uniformly-weighted mixture model. In other words, we combine the predictions as $\rho(t|\mathbf{x}) = K^{-1}\sum_{k=1}^K\rho_{\theta_k}(t|\mathbf{x},\theta_k)$. Letting the predictive distributions of the mixture be Gaussian, $K^{-1}\sum_{k=1}^K \mathcal{N}\big(\mu_{\theta_k}(\mathbf{x}),\sigma^2_{\theta_k}(\mathbf{x})\big)$, the resulting statistically integrated mean and variance are given by $\mu_*(\mathbf{x}) = K^{-1}\sum_{k=1}^K\mu_{\theta_k}(\mathbf{x})$ and $\sigma^2_*(\mathbf{x}) = K^{-1}\sum_{k=1}^K\big(\sigma^2_{\theta_k}(\mathbf{x}) + \mu_{\theta_k}(\mathbf{x})\big) - \mu_*^2(\mathbf{x})$, respectively. These quantities are evaluated against the test set. It is worth noting that, in all our experiments, we \mbox{train an ensemble of five NNs (i.e. $K = 5$).}



\section{Experimental results}
\label{experimentalresults}
We evaluate our proposed method both qualitatively and quantitatively through a series of experiments on regression benchmark datasets. In particular, we first conduct a regression experiment on a one-dimensional toy dataset, and then experiment with well-known, real world datasets.\footnote{The datasets can be found at the University of California, Irvine (UCI) machine learning data repository.} Further, we show that our approach outperforms state-of-the-art methods in out-of-distribution generalization. In our experiments, we train NNs with one hidden layer and use the $\epsilon$-BVM loss function $\mathcal{C}_{\text{BVM}}\big(B(\epsilon)\big)$ described by Equation (\ref{bvmcdf}) (in what follows, we will refer to the $\epsilon$-BVM loss as simply the BVM loss). We randomly initialize the NN weights (using the PyTorch default weight initialization) and randomly shuffle the paired training examples. 

\subsection{Toy dataset}
We first qualitatively assess the performance of our proposed method on a toy dataset that was used in \cite{hernandez2015, lakshminarayanan2017}. The dataset is produced by uniformly sampling (at random) $20$ inputs $x$ in the interval $[-4, 4]$. The label $t$ corresponding to each input $x$ is obtained by computing $t = x^3 + \xi$ where $\xi \sim \mathcal{N}(0, 3^2)$. The NN architecture consists of one layer with 100 hidden units and the value of $\epsilon$ in the BVM loss is set to $1$ as the data is not normalized.\\

\begin{figure*}[t]
\centering
\begin{subfigure}{5.35cm}
\includegraphics[width=5.35cm]{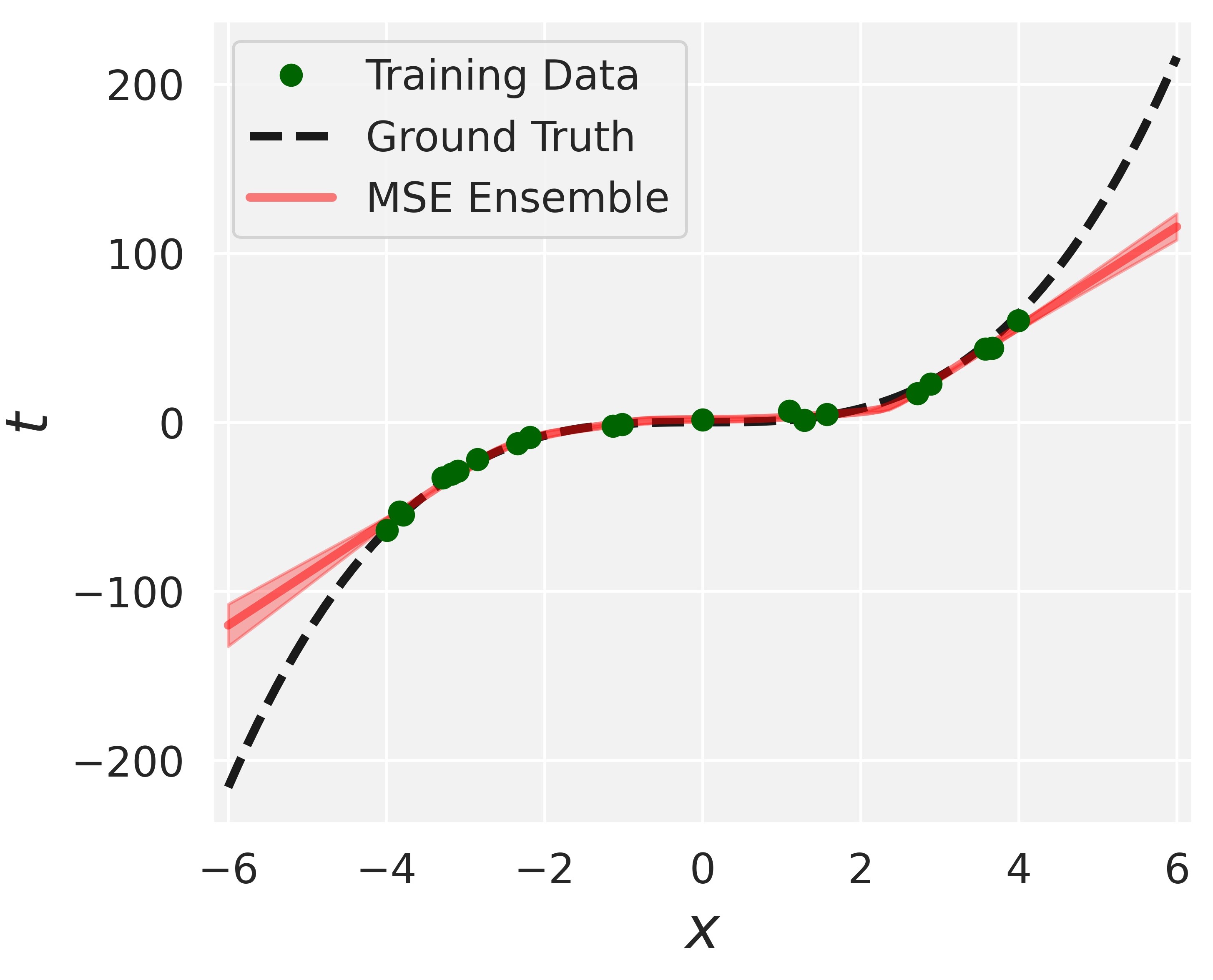}
\caption{Ensemble-5 (MSE).\label{fig1a}}
\end{subfigure}
\:\:
\begin{subfigure}{5.05cm}
\includegraphics[width=5.05cm]{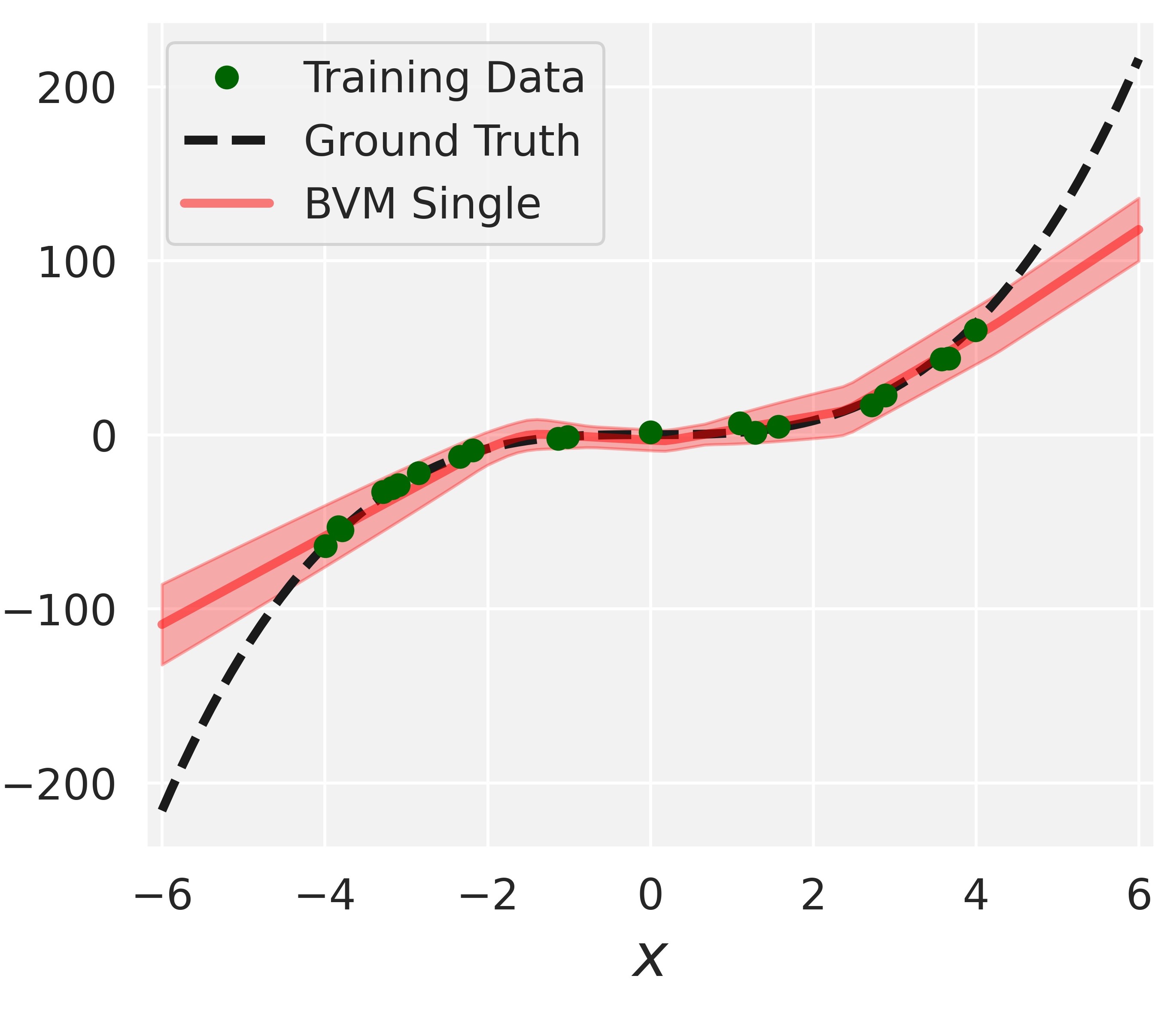}
\caption{Single (BVM).\label{fig1b}}
\end{subfigure}
\:\:
\begin{subfigure}{5.05cm}
\includegraphics[width=5.05cm]{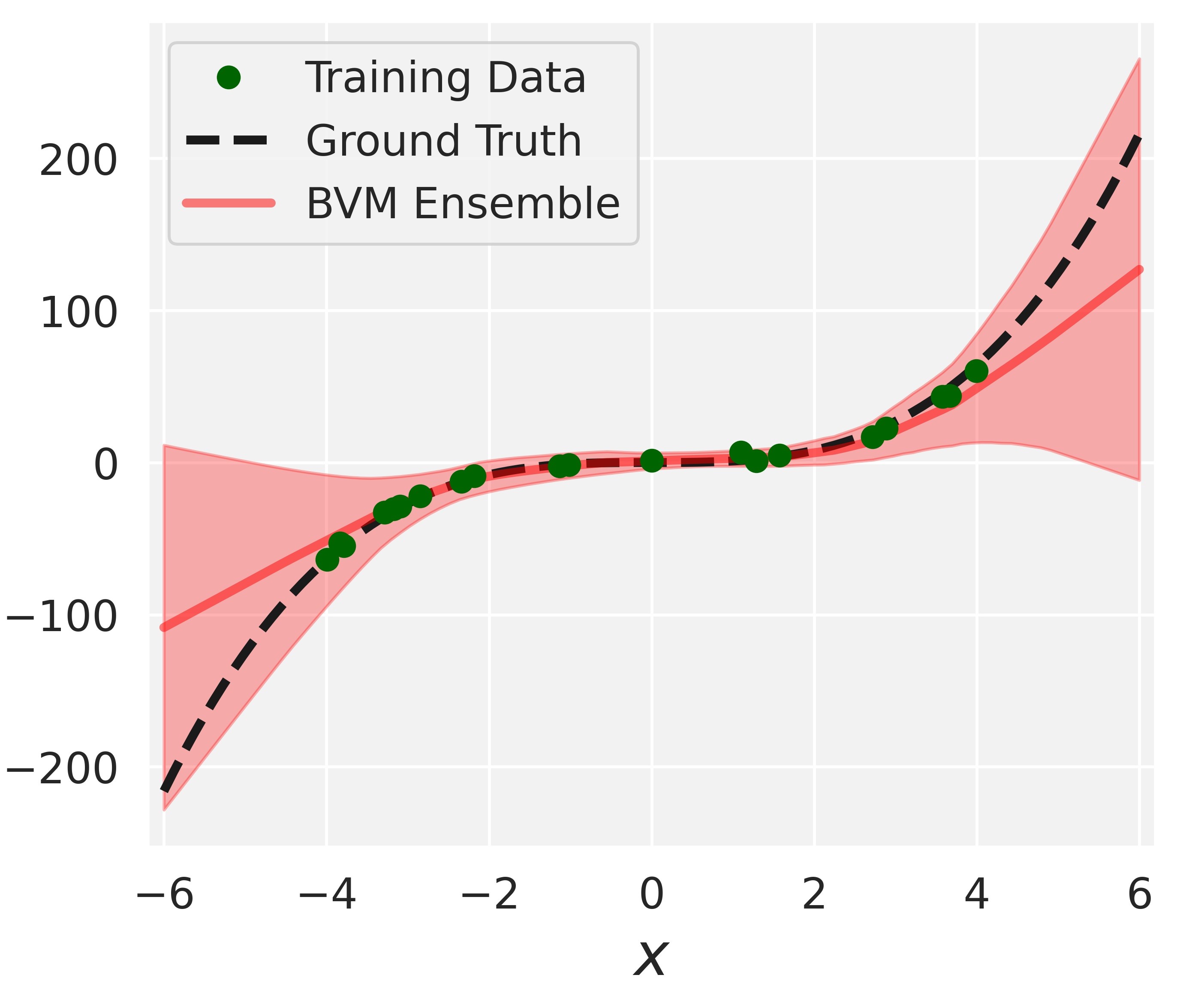}
\caption{Enemble-5 (BVM).\label{fig1c}}
\end{subfigure}
\caption{Regression on a toy dataset. The pink shaded area corresponds to $\mu \pm 3\hspace{0.8pt}\sigma$. Figure \ref{fig1a} corresponds to the variance of $5$ networks trained independently using MSE, Figure \ref{fig1b} corresponds to training a single network using BVM, and Figure \ref{fig1c} corresponds to training an ensemble of $5$ networks using BVM. \label{fig1}}
\end{figure*}

In order to measure and estimate uncertainty, a commonly used approach is to train multiple NNs independently (i.e. an ensemble of NNs) to minimize MSE, and compute the variance of the networks' generated point predictions. We show that learning the variance by training using the BVM loss function results in better predictive uncertainty estimation. The results are shown in Figure \ref{fig1}.\\

From Figure \ref{fig1}, it is clear that predictive uncertainty estimation can be improved by learning the variance through training using the BVM loss, and it can be further improved by training an ensemble of NNs (the effect of ensemble learning becomes more apparent as we move further away from the training data). Note that the results we get using the proposed BVM loss are very similar to the results produced using NLL in \cite{lakshminarayanan2017} since $\epsilon$ is small relative to the range of the data. The goal of this experiment is to show that the BVM loss function is indeed suitable for predictive uncertainty estimation by reproducing the results in \cite{lakshminarayanan2017}.

\subsection{Training using MSE vs NLL vs BVM}
This section shows that the predicted variance (using our method) is as well-calibrated as the one from Deep Ensembles (using NLL) and is better calibrated than the empirical variance (using MSE). In \cite{lakshminarayanan2017}, it was shown that training an ensemble of NNs with a single output (representing the mean) using MSE and computing the empirical variance of the networks’ predictions to estimate uncertainty does not lead to well-calibrated predictive probabilities. This was due to the fact that MSE does not capture predictive uncertainty. It was then shown that learning the predictive variance by training an ensemble of NNs with two outputs (corresponding to the mean and variance) using NLL (i.e. Deep Ensembles) results in well-calibrated predictions. We show that this is also the case for the proposed BVM loss.\\

We reproduce an experiment from \cite{lakshminarayanan2017} using the BVM loss function (with $\epsilon = 0.01$), where we construct reliability diagrams (also known as calibration curves) on the benchmark datasets. The procedure is as follows: (i) we calculate the $z\%$ prediction interval for each test point (using the predicted mean and variance), (ii) we then measure the actual fraction of test observations that fall within this prediction interval,
and (iii) we repeat the calculations for $z = 10\%,\hdots, 90\%$ in steps of $10$. 
If the actual fraction is close to the expected fraction (i.e. $\approx z\%$), this indicates that the predictive probabilities are well-calibrated. The ideal output would be a diagonal line. In other words, a regressor is considered to be well-calibrated if its calibration curve is close to the diagonal.\\

\begin{figure}[h]
\centering
\includegraphics[width=9cm]{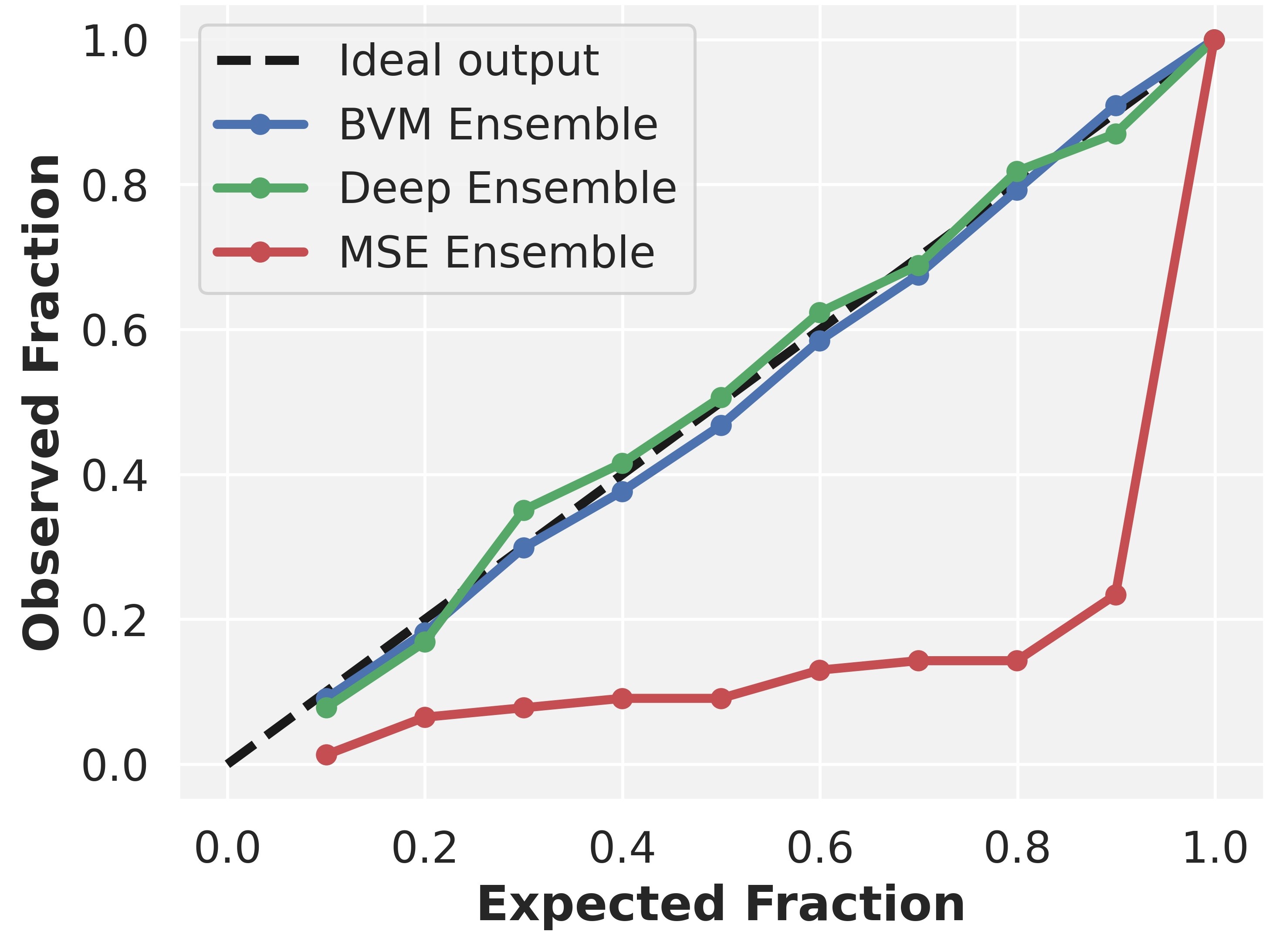}
\caption{Reliability diagram for the \emph{Energy} dataset. The predicted variance using our approach is as well-calibrated as the one from Deep Ensembles (using NLL) and is better calibrated than the empirical variance using MSE, which is overconfident.\label{fig2}}
\end{figure}
\noindent\vspace{-15pt}\\\indent
We report the reliability diagram for the \emph{Energy} dataset in Figure \ref{fig2}; diagrams for the other regression benchmark datasets are reported in \ref{AppendixA} (the trend is the same for all datasets). We find that our method provides well-calibrated uncertainty estimates with a calibration curve very close to the diagonal (and almost overlapping with the curve of Deep \mbox{Ensembles \cite{lakshminarayanan2017}).} We also find that the predicted variance (learned using BVM or NLL) is better calibrated than the empirical variance (computed by training five NNs using MSE) which is overconfident. For instance, for the $40\%$ prediction interval (i.e. the expected fraction is equal to $0.4$), the actual fraction of test observations that fall within the interval is only $10\%$ (i.e. the observed fraction is around $0.1$). In other words, the empirical variance (using MSE) underestimates the true uncertainty. We also report the Expected Calibration Error (ECE) and the Maximum Calibration Error (MCE) \cite{naeini2015obtaining, guo2017calibration} corresponding to MSE Ensemble, Deep Ensemble, and BVM Ensemble on the regression benchmarks datasets. The results agree with those in Figure \ref{fig2} and are reported in \ref{AppendixA}. 

\subsection{Real world datasets}
\label{realworlddatasets}

We further evaluate our proposed method by comparing it to existing state-of-the-art methods. We adopt the same experimental setup as in \cite{hernandez2015} for evaluating PBP, \cite{gal2016} for evaluating MC-dropout, and \cite{lakshminarayanan2017} for evaluating Deep Ensembles. We use one-hidden-layer NNs with Rectified Linear Unit (ReLU) activation function \cite{nair2010}, consisting of $50$ hidden units for all datasets except for the largest one (i.e. \textit{Protein}) where we use NNs with $100$ hidden units. We train NNs using the BVM loss function with $\epsilon = 0.01$. Each dataset is randomly split into training and test sets with $90\%$ and $10\%$ of the available data, respectively. For each train-test split, we train an ensemble of $5$ networks. We repeat the splitting process $20$ times and report the average test performance of our proposed method. For the larger \textit{Protein} dataset, we perform the train-test splitting $5$ times (instead of 20).\\

In our experiments, we run the training for $40$ epochs, using mini-batches of size $32$ and AdamW optimizer with fixed learning rate of $3\times 10^{-4}$. For all the datasets, we apply feature scaling by standardizing the input features to have zero mean and unit variance, and normalize the targets to have a range of $[0, 1]$ (in the training set). Before evaluating the predictions, we invert the normalization factor on the predictions so they are back to the original scale of the targets for the purpose of error evaluation. Note that a sigmoid activation function is applied to the outputs of the NNs corresponding to the mean and variance. We summarize our results in Table \ref{table1}, along with the results of PBP, MC-dropout, and Deep Ensembles as were outlined in their respective papers. For each dataset, the best method(s) is (are) highlighted in bold.\\

The results in Table \ref{table1} clearly demonstrate that our proposed method is competitive with existing state-of-the-art \mbox{methods}. As might be expected, our method performs sub-optimally compared to other methods in terms of RMSE (e.g. on the \emph{Energy} dataset). Since our method optimizes for the BVM loss, which learns both the mean and the variance (to better capture uncertainties) rather than learning only the mean, it gives less optimal RMSE values. Also note that, although our method outperforms PBP and MC-dropout in terms of NLL on many datasets, it did not outperform Deep Ensembles (e.g. on the \emph{Energy} dataset, our method produces the second lowest NLL average of $1.67$ behind Deep Ensembles whose NLL average is $1.38$). Since the Deep Ensembles method optimizes for NLL, it is expected to perform better than the BVM approach for $\epsilon>0$ -- \emph{at least when the splitting of the data into training and test sets is done randomly} (i.e. when tested on in-distribution data). The methods are comparable and identical in the limit $\epsilon \rightarrow 0$, because the BVM loss becomes equivalent to NLL. We \mbox{intentionally} used a nonzero $\epsilon$ to highlight its effect on the predictions (compared to Deep Ensembles) when tested on in-distribution samples. We later introduce and apply the concept of ``outlier train-test splitting'', and show that our method outperforms Deep Ensembles when evaluated on out-of-distribution samples (see Section \ref{oodgen}). 

\begin{table*}[t]
\centering
\caption{Average test performance in RMSE and NLL on regression benchmark datasets.\label{table1}}
\scalebox{.53}{
\begin{tabular}{lrr|cccc|cccc}
\toprule
& & &\multicolumn{4}{ c| }{Avg. Test RMSE and Std. Errors} & \multicolumn{4}{ c }{Avg. Test NLL and Std. Errors}\\
\hhline{~~~~~~~}
\textbf{Dataset}& $N$ & $d$ & \textbf{PBP} & \textbf{MC-dropout} & \textbf{Deep Ensembles} & \textbf{BVM} & \textbf{PBP} & \textbf{MC-dropout} & \textbf{Deep Ensembles} & \textbf{BVM} \\
\midrule
Boston housing & 506 & 13 & \textbf{3.01} $\pm$ \textbf{0.18} & \textbf{2.97} $\pm$ \textbf{0.19} & \textbf{3.28} $\pm$ \textbf{1.00} & \textbf{3.06} $\pm$ \textbf{0.22} & \textbf{2.57} $\pm$ \textbf{0.09} & \textbf{2.46} $\pm$ \textbf{0.06} & \textbf{2.41} $\pm$ \textbf{0.25} & \textbf{2.52} $\pm$ \textbf{0.08} \\
Concrete & 1,030 & 8 & 5.67 $\pm$ 0.09 & \textbf{5.23} $\pm$ \textbf{0.12} & 6.03 $\pm$ 0.58 & 6.07 $\pm$ 0.18 & \textbf{3.16} $\pm$ \textbf{0.02} & \textbf{3.04} $\pm$ \textbf{0.02} & \textbf{3.06} $\pm$ \textbf{0.18} & \textbf{3.18} $\pm$ \textbf{0.14}\\
Energy & 768 & 8 & 1.80 $\pm$ 0.05 & \textbf{1.66} $\pm$ \textbf{0.04} & 2.09 $\pm$ 0.29 & 2.16 $\pm$ 0.07 & 2.04 $\pm$ 0.02 & 1.99 $\pm$ 0.02 & \textbf{1.38} $\pm$ \textbf{0.22} & \textbf{1.67} $\pm$ \textbf{0.13}\\
Kin8nm & 8,192 & 8 & 0.10 $\pm$ 0.00 & 0.10 $\pm$ 0.00 & \textbf{0.09} $\pm$ \textbf{0.00} & 0.11 $\pm$ 0.00 & -0.90 $\pm$ 0.01 & -0.95 $\pm$ 0.01 & \textbf{-1.20} $\pm$ \textbf{0.02} & -0.85 $\pm$ 0.10\\
Naval propulsion plant & 11,934 & 16 & 0.01 $\pm$ 0.00 & 0.01 $\pm$ 0.00 & \textbf{0.00} $\pm$ \textbf{0.00} & 0.01 $\pm$ 0.00 & -3.73 $\pm$ 0.01 & -3.80 $\pm$ 0.01 & \textbf{-5.63} $\pm$ \textbf{0.05} & -3.92 $\pm$ 0.01\\
Power plant & 9,568 & 4 & 4.12 $\pm$ 0.03 & \textbf{4.02} $\pm$ \textbf{0.04} & \textbf{4.11} $\pm$ \textbf{0.17} & \textbf{4.18} $\pm$ \textbf{0.13} & 2.84 $\pm$ 0.01 & \textbf{2.80} $\pm$ \textbf{0.01} & \textbf{2.79} $\pm$ \textbf{0.04} & 3.07 $\pm$ 0.08\\
Protein & 45,730 & 9 & 4.73 $\pm$ 0.01 & \textbf{4.36} $\pm$ \textbf{0.01} & 4.71 $\pm$ 0.06 & \textbf{4.29} $\pm$ \textbf{0.08} & 2.97 $\pm$ 0.00 & 2.89 $\pm$ 0.00 & \textbf{2.83} $\pm$ \textbf{0.02} & 3.02 $\pm$ 0.03\\
Wine & 1,599 & 11 & \textbf{0.64} $\pm$ \textbf{0.01} & \textbf{0.62} $\pm$ \textbf{0.01} & \textbf{0.64} $\pm$ \textbf{0.04} & \textbf{0.64} $\pm$ \textbf{0.01} & 0.97 $\pm$ 0.01 & \textbf{0.93} $\pm$ \textbf{0.01} & \textbf{0.94} $\pm$ \textbf{0.12} & \textbf{1.01} $\pm$ \textbf{0.09}\\
Yacht & 308 & 6 & \textbf{1.02} $\pm$ \textbf{0.05} & \textbf{1.11} $\pm$ \textbf{0.09} & 1.58 $\pm$ 0.48 & 1.67 $\pm$ 0.25 & 1.63 $\pm$ 0.02 & 1.55 $\pm$ 0.03 & \textbf{1.18} $\pm$ \textbf{0.21} & \textbf{1.56} $\pm$ \textbf{0.18}\\
\bottomrule
\end{tabular}
}
\end{table*}
\noindent\vspace{-18pt}\\
\subsection{Robustness and out-of-distribution generalization}
\label{oodgen}
We aim to show that our proposed method is robust and able to generalize better to out-of-distribution (OOD) data than Deep Ensembles. That is, if we evaluate our method on data that is statistically different from the training data, we observe more robustness and higher predictive uncertainties.

\subsubsection{Experiment 1}

\begin{figure}[t]
\centering
\includegraphics[width=9cm]{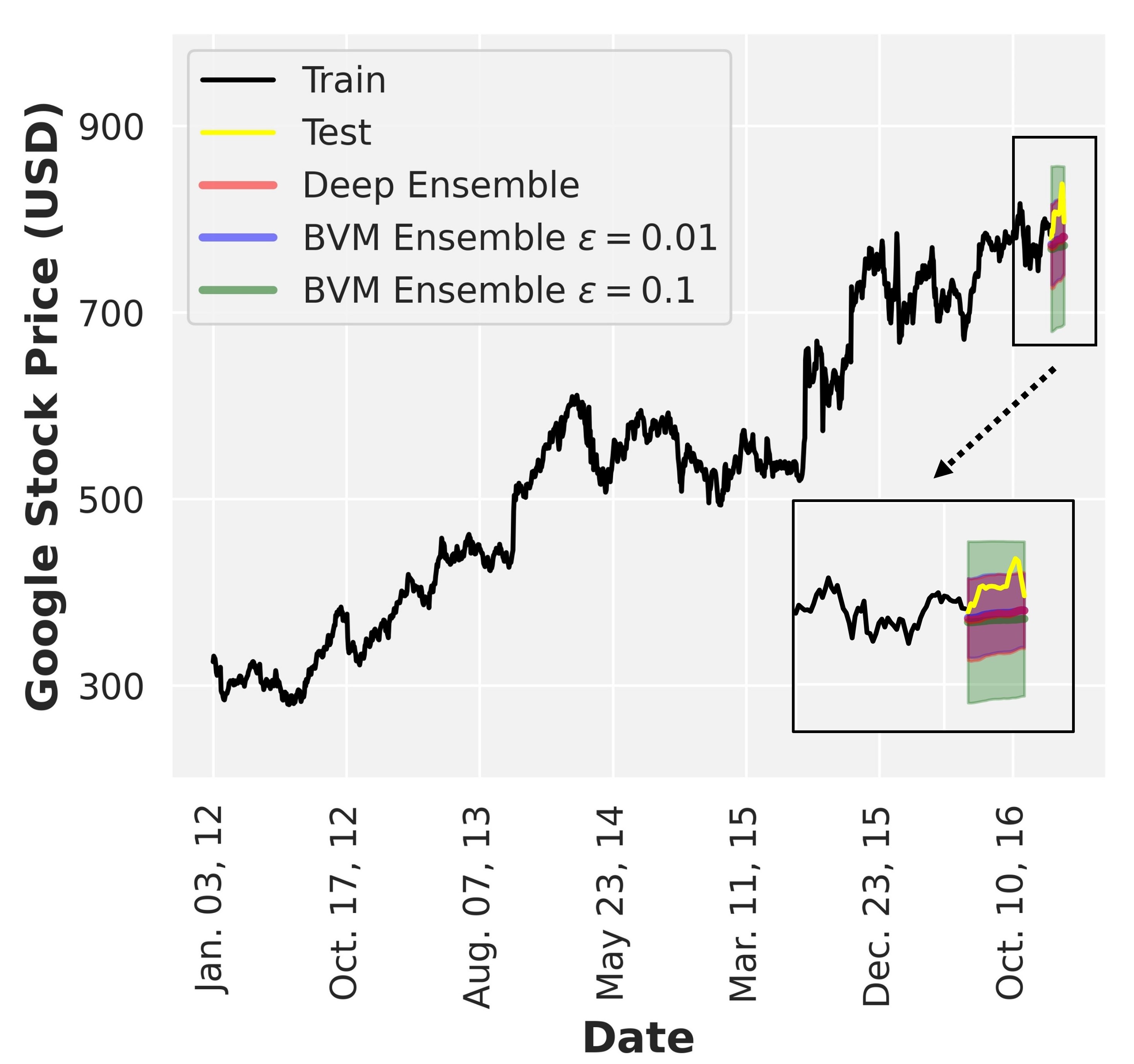}
\caption{Google stock price predictive response and uncertainty: Note that using BVM Ensemble with $\epsilon = 0.01$ results in a predictive envelope (blue) that almost overlaps with the predictive envelope of Deep Ensembles (red). By increasing the value of $\epsilon$ to $0.1$ in the BVM loss, all the test points (yellow) fall within the predictive envelope (green).\label{fig4}}
\end{figure}

We consider a training set consisting of Google stock prices for a period of $5$ years (from the beginning of $2012$ till the end of $2016$) and a test set containing the stock prices of January $2017$ (see Figure \ref{fig4}). In particular, we consider the Google opening stock price, i.e. the stock price at the beginning of the financial/trading day. It is worth noting that the input feature vector is $60$-dimensional corresponding to a $60$-day window, i.e. for a given day, the NN will consider the stock prices for the past $60$ days, and based on the trends captured during this time window, it will predict the corresponding stock price (with its uncertainty).\\

We train an ensemble of $5$ NNs consisting of $4$ hidden layers with $50$ hidden units per layer.\footnote{Indeed, training recurrent NNs will improve forecasting accuracies, however, here we are more interested in predictive uncertainties and standard NNs were enough to prove our point.} We run the training for $40$ epochs, using batch size of $32$ and Adam optimizer with fixed learning rate of $1\times 10^{-3}$. We repeat this process for three different loss functions: (i) The NLL loss in (\ref{nll}) used in Deep Ensembles \cite{lakshminarayanan2017}, (ii) the BVM loss in (\ref{bvmcdf}) with $\epsilon = 0.01$, and (iii) the BVM loss with $\epsilon = 0.1$. We plot the predicted mean stock price along with the $95\%$ prediction interval corresponding to January $2017$. The results \mbox{are shown in Figure \ref{fig4}.}\\

The results clearly demonstrate that the value of $\epsilon$ in the BVM loss affects the predictive uncertainty (i.e. the prediction interval). A small value of $\epsilon$ corresponds to a stricter agreement condition between the NN predictive means and the observed targets, which results in a narrower prediction interval (i.e. lower variance values). Note that using $\epsilon = 0.01$ results in a predictive envelope that almost overlaps with the prediction interval of Deep Ensembles. When we increase the value of $\epsilon$ to $0.1$ in the BVM loss, the agreement conditions become less stringent, and this leads to a wider prediction interval, which better captures the uncertainty of the stock price in the test set. This results in a lower NLL for \mbox{this highly volatile test set.}\\

\begin{table}[h]
\centering
\caption{Test performance in NLL on Google stocks dataset.\label{table2}}
\scalebox{.75}{
\begin{tabular}{lrr|ccc}
\toprule
& & & \multicolumn{3}{ c }{Test NLL}\\
\hhline{~~~~~~}
\textbf{Dataset}& $N$ & $d$ & \textbf{Deep Ensembles} & \textbf{BVM} $\boldsymbol{(\epsilon = 0.01)}$ & \textbf{BVM} $\boldsymbol{(\epsilon = 0.1)}$ \\
\midrule
Google stocks & 1,198 & 60 & 5.23 & 5.19 & \textbf{5.13} \\
\bottomrule
\end{tabular}
}
\end{table}

The NLL results are summarized in Table \ref{table2}. Due to the out-of-distribution nature of the test data, the BVM loss with a relatively large $\epsilon = 0.1$ results in the lowest NLL. Since this loss leads to the largest variances (or uncertainties), its corresponding likelihood (\ref{maxlikelihood}) will be the largest, which is equivalent to the lowest NLL. Using a very large $\epsilon$ will overly coarse grain the data and one will \mbox{lose predictive power.} 



\subsubsection{Experiment 2}
We now compare our method to Deep Ensembles in terms of robustness and OOD generalization on the regression benchmark datasets used in Section \ref{realworlddatasets} (see table \ref{table1}). Since the presence of OOD examples (for testing) is not guaranteed in these datasets, we apply ``outlier train-test splitting", which forces the generation of statistical differences between the training and test sets (when outliers exist). We repeat the experiment on the benchmarks from Section \ref{realworlddatasets}, however, instead of randomly splitting the datasets into training and test sets, we now detect outliers (e.g. $10\%$ of the dataset) and treat them as test examples and train on the remaining examples (e.g. $90\%$ of the dataset). The outliers represent out-of-distribution examples that could potentially lead to heavy losses if characterized poorly in a deployment environment. Using this splitting process, we can better evaluate a method's predictive ability on out-of-distribution samples. To perform the outlier train-test data splitting, we use Isolation Forest \cite{liu2008isolation} to detect the outliers in the datasets (which isolates anomalies that are less frequent and different \mbox{in the feature space).}\\

We train an ensemble of $5$ NNs consisting of one hidden layer with $50$ hidden units, using both the NLL loss (which is Deep Ensembles) and the BVM loss function (with $\epsilon = 0.01$). We train for $40$ epochs, with batch size of $16$ and Adam optimizer with fixed learning rate of $3\times 10^{-3}$. We summarize our results along with the statistical differences between the normalized train-test targets in Table \ref{table3}. For each dataset, the best method is highlighted in bold.\\


As shown in Table \ref{table3}, our method consistently outperforms Deep Ensembles on all datasets that have significant statistical differences between the training and test sets (gray rows). In other words, the BVM approach is robust to statistical change. It is interesting to note that while our method did not directly optimize for NLL, it was still able to outperform Deep Ensembles, which did.
\clearpage
\begin{table*}[t]
\centering
\caption{Test performance in NLL on top $10\%$ outliers in regression benchmark datasets, along with the statistical differences between the normalized train-test targets. The datasets with large statistical differences are highlighted in gray.\label{table3}}
\scalebox{.75}{
\begin{tabular}{lrr|cc|cc}
\toprule
& & &\multicolumn{2}{ c| }{Statistical Difference} & \multicolumn{2}{ c }{Test NLL}\\
\hhline{~~~~~~~}
\textbf{Dataset}& $N$ & $d$ & $\boldsymbol{\mu}\textbf{\big(}\boldsymbol{t}_\textbf{test}\textbf{\big)} - \boldsymbol{\mu}\textbf{\big(}\boldsymbol{t}_\textbf{train}\textbf{\big)}$ & $\boldsymbol{\sigma^2}\textbf{\big(}\boldsymbol{t}_\textbf{test}\textbf{\big)} - \boldsymbol{\sigma^2}\textbf{\big(}\boldsymbol{t}_\textbf{train}\textbf{\big)}$ & \textbf{Deep Ensembles} & \textbf{BVM} \\
\midrule
Boston housing & 506 & 13 & \cellcolor{gray!25}\phantom{-\,}0.05 & \cellcolor{gray!25}\phantom{-\,}0.06 & \cellcolor{gray!25}\phantom{-\,\,}4.51 &\cellcolor{gray!25} \phantom{-}\textbf{3.92} \\
Concrete & 1,030 & 8 & \cellcolor{gray!25}\phantom{-\,}0.14 &\cellcolor{gray!25} -0.01 &\cellcolor{gray!25} \phantom{-}4.12 & \cellcolor{gray!25}\phantom{-\,}\textbf{3.84} \\
Energy & 768 & 8 & \cellcolor{gray!25}\phantom{-\,}0.06 &\cellcolor{gray!25} \phantom{-}0.01 &\cellcolor{gray!25} \phantom{-}2.98 & \cellcolor{gray!25}\phantom{-\,}\textbf{2.57}\\
Kin8nm & 8,192 & 8 &\cellcolor{gray!25} -0.04 &\cellcolor{gray!25} \phantom{-}0.01 &\cellcolor{gray!25} -0.85 & \cellcolor{gray!25}\phantom{\,}\textbf{-0.87}\\
Naval propulsion plant & 11,934 & 16 & \phantom{-\,\,}0.01 & \phantom{-\,\,}0.02 & \phantom{-}\textbf{-4.42} & \phantom{\,}-3.84 \\
Power plant & 9,568 & 4 & \phantom{-}-0.01 & \phantom{-\,\,}0.02 & \phantom{-\,\,}\textbf{2.82} & \phantom{-\,}3.18\\
Protein & 45,730 & 9 & \phantom{-\,\,}0.00 & \phantom{-\,\,}0.01 & \phantom{-\,\,}\textbf{2.86} & \phantom{-\,}3.09 \\
Wine & 1,599 & 11 & \cellcolor{gray!25} \phantom{-}0.04 & \cellcolor{gray!25} \phantom{-}0.02 & \cellcolor{gray!25} \phantom{-}3.15 & \cellcolor{gray!25} \phantom{-}\textbf{1.47} \\
Yacht & 308 & 6 & \cellcolor{gray!25} \phantom{-}0.26 & \cellcolor{gray!25} \phantom{-}0.10 &\cellcolor{gray!25} \phantom{-}3.95 &\cellcolor{gray!25} \phantom{-}\textbf{1.83} \\
\bottomrule
\end{tabular}
}
\end{table*}

\allowdisplaybreaks
\subsubsection{Why does BVM outperform Deep Ensembles on OOD samples?\label{oodsec}}
Note that for a given input feature vector $\mathbf{x}_n$, the minimizer of the BVM loss function satisfies\vspace{-5pt}\\
\begin{align}
&\underset{\mu_n,\sigma_n}{\text{argmin}}\,-\log\Bigg[\Phi\bigg(\frac{t_n+\epsilon-\mu_n}{\sigma_n}\bigg) - \Phi\bigg(\frac{t_n-\epsilon-\mu_n}{\sigma_n}\bigg)\Bigg]\notag\\[\medskipamount]
=\,\, &\underset{\mu_n,\sigma_n}{\text{argmin}}\,-\log\Bigg(\frac{1}{2\epsilon}\bigg[\Phi\bigg(\frac{t_n+\epsilon-\mu_n}{\sigma_n}\bigg) - \Phi\bigg(\frac{t_n-\epsilon-\mu_n}{\sigma_n}\bigg)\bigg]\Bigg)
%
\end{align}
Taylor expanding around $\epsilon = 0$ leads to
\begin{align}
-&\log\Bigg(\frac{1}{2\epsilon}\bigg[\Phi\bigg(\frac{t_n+\epsilon-\mu_n}{\sigma_n}\bigg) - \Phi\bigg(\frac{t_n-\epsilon-\mu_n}{\sigma_n}\bigg)\bigg]\Bigg)\notag\\[\medskipamount]
\simeq & \frac{1}{2}\log 2\pi\sigma^2_n + \frac{\big(t_n-\mu_n\big)^2}{2\sigma^2_n} - \frac{\epsilon^2}{6}\,\Bigg[\frac{\big(t_n-\mu_n\big)^2}{\sigma^4_n}-\frac{1}{\sigma_n^2}\Bigg] + \mathcal{O}(\epsilon^3)
\end{align}
The proof can be found in \ref{AppendixB}. Thus, the minimizer of the BVM loss over the set of all input feature vectors can be approximated as
\begin{align}
\underset{\mu_n,\sigma_n}{\text{argmin}}\,\underbrace{\frac{1}{N}\sum_{n=1}^N\Bigg(\frac{1}{2}\log 2\pi\sigma^2_n + \frac{\big(t_n-\mu_n\big)^2}{2\sigma^2_n}}_{\mathcal{C}_{\text{NLL}}} - \frac{\epsilon^2}{6}\,\Bigg[\frac{\big(t_n-\mu_n\big)^2}{\sigma^4_n}-\frac{1}{\sigma_n^2}\Bigg]\Bigg)
\end{align}
We can clearly see that for $\epsilon = 0$, the minimizer of the BVM loss is indeed the minimizer of the NLL loss in Equation (\ref{nll}). For a nonzero $\epsilon$, $\sigma_n$ will increase linearly with $\epsilon$ (see proof in \ref{AppendixB}) leading to a larger variance (as in \cite{levi2019evaluating, laves2021recalibration}), and hence a wider distribution (or prediction interval). Thus, the OOD samples near the tails (i.e. the outliers) will be more probable resulting in lower NLL values compared to Deep Ensembles (keeping in mind that the in-distribution samples near the mean will be less probable resulting in higher NLL values compared to Deep Ensembles, which was \mbox{the case in Table \ref{table1}).}

\section{Discussion}
The numerical experiments above show that the results depend on the value of $\epsilon$. Hence, some guidance is provided on how to select $\epsilon$. One approach is to treat $\epsilon$ as a parameter to be learned during the training process. Another approach is to treat $\epsilon$ as a hyperparameter and perform a grid search for its optimal value on a validation set.\\

While post-hoc calibration methods such as \cite{guo2017calibration, levi2019evaluating, laves2021recalibration} provide the benefit of optimizing for the scaling parameter on a calibration set after training, finding $\epsilon$ that yield optimal calibration is computationally demanding (e.g. using Bayesian optimization), as $\epsilon$ is chosen prior to (or learned during) training. In this case, one can adopt the recently proposed post-hoc calibration method \cite{rahimi2020post} which suggests training additional NNs' layers on the calibration set. In other words, optimal calibration is achieved by training the parameters of the additional layers on the calibration set rather than optimizing for $\epsilon$.

\section{Conclusion}

In this work, we proposed a new loss function for regression uncertainty estimation (based on the BVM framework) which reproduces maximum likelihood estimation in the limiting case. This loss, boosted by ensemble learning, improves predictive performance when the training and test sets are statistically different. Experiments on in-distribution data show that our method generates well-calibrated uncertainty estimates and is competitive with existing state-of-the-art methods. When tested on out-of-distribution samples (outliers), our method exhibits superior predictive power by consistently displaying improved predictive log-likelihoods. Because the data source statistics in the learning and deployed environments are often known to be different, our method can be used to improve decision-making in the deployed environment, which is generally critical to the success and advancement of applications in reliability engineering and system safety. Although we focus in the article on one particular loss function, the BVM framework is versatile and general enough to allow for the derivation of other loss functions that can be used for predictive uncertainty estimation. However, some BVM-based loss functions may not have an elegant closed-form expression as the one presented in the article, and hence, they may be more computationally expensive. Our future work involves expanding the BVM framework to address predictive uncertainty estimation in classification and vision problems, and its real-world deployment in safety-critical systems and applications.




\paragraph{\textbf{Acknowledgments}}
This work was supported by the Center for Complex Systems (CCS) at King Abdulaziz City for Science and Technology (KACST) and the Massachusetts Institute of Technology (MIT). We would like to thank all the researchers in the CCS. The first author was also supported by the Mathworks Engineering Fellowship.

\bibliography{mybibfile}
\clearpage
\appendix
\section{Training using MSE vs NLL vs BVM}
\label{AppendixA}
\vspace*{3pt}
This section shows that the predicted variance (using our method) is as well-calibrated as the one from Deep Ensembles (using NLL) and is better calibrated than the empirical variance (using MSE). In \cite{lakshminarayanan2017}, it was shown that training an ensemble of NNs with a single output (representing the mean) using MSE and computing the empirical variance of the networks’ predictions to estimate uncertainty does not lead to well-calibrated predictive probabilities. This was due to the fact that MSE does not capture predictive uncertainty. It was then shown that learning the predictive variance by training NNs with two outputs (corresponding to the mean and variance) using NLL (i.e. Deep Ensembles) results in well-calibrated predictions. We show that this is also the case for the proposed BVM loss.\\

We reproduce an experiment from \cite{lakshminarayanan2017} using the BVM loss function, where we construct reliability diagrams (also known as calibration curves) on the benchmark datasets. The procedure is as follows: (i) we calculate the $z\%$ prediction interval for each test point (using the predicted mean and variance), (ii) we then measure the actual fraction of test observations that fall within this prediction interval,
and (iii) we repeat the calculations for $z = 10\%,\hdots, 90\%$ in steps of $10$. 
If the actual fraction is close to the expected fraction (i.e. $\approx z\%$), this indicates that the predictive probabilities are well-calibrated. The ideal output would be the diagonal line. In other words, a regressor is considered to be well-calibrated if its calibration curve is close to the diagonal.\\

\begin{figure}[h]
\centering
\includegraphics[width=15cm]{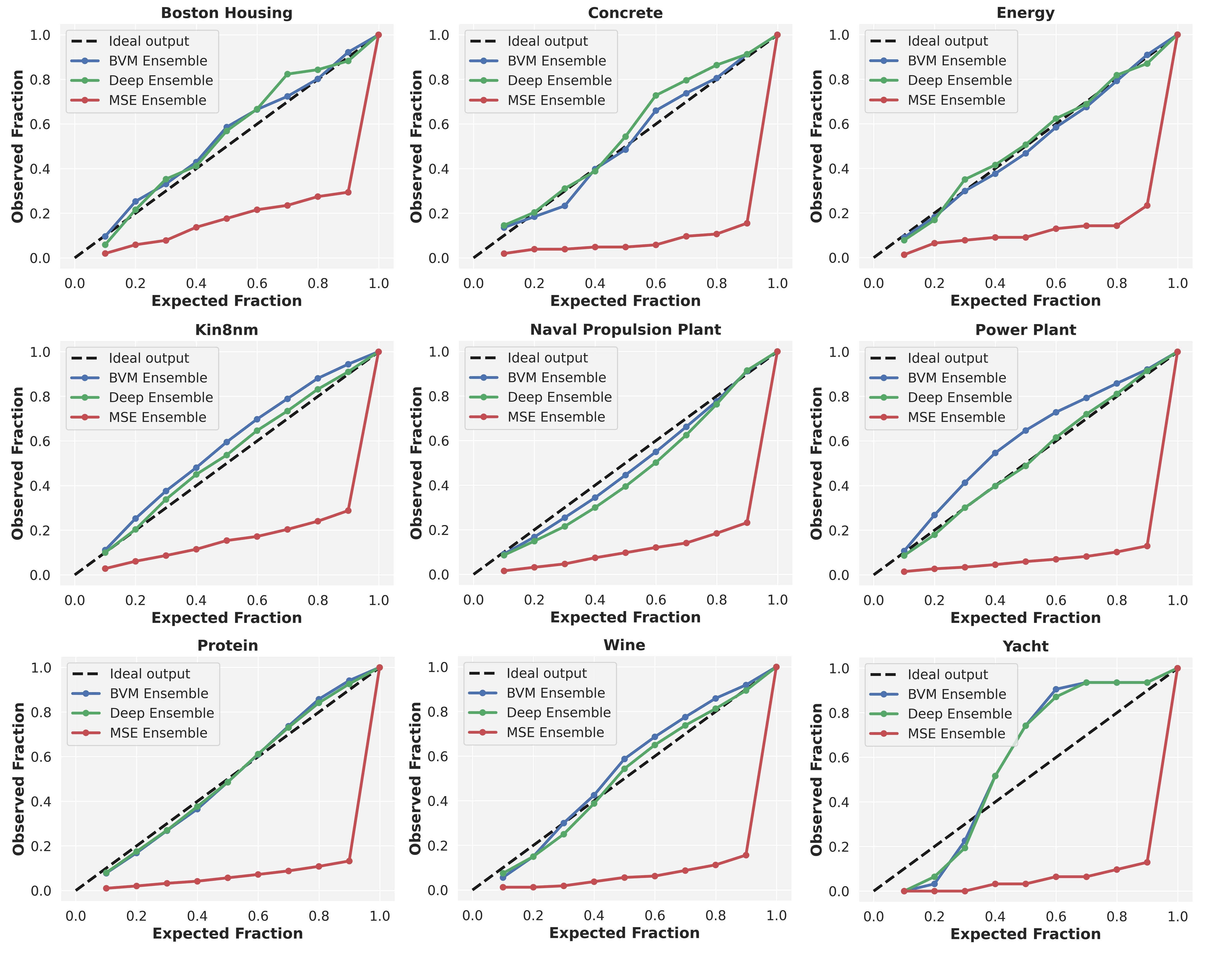}
\caption{Reliability diagrams for the benchmark datasets. The predicted variance using our approach is as well-calibrated as the one from Deep Ensembles (using NLL) and is better calibrated than the empirical variance using MSE. \label{figreldiag}}
\end{figure}

We report the reliability diagrams for the benchmark datasets in Figure \ref{figreldiag}. We find that our method provides well-calibrated uncertainty estimates with a calibration curve very close to the diagonal (and almost overlapping with the curve of Deep Ensembles \cite{lakshminarayanan2017}). We also find that the predicted variance (learned using BVM or NLL) is better calibrated than the empirical variance (computed by training five NNs using MSE) which is overconfident. For instance, if we consider the reliability diagram for the \emph{Boston Housing} dataset, for the $60\%$ prediction interval (i.e. the expected fraction is equal to $0.6$), the actual fraction of test observations that fall within the interval is only $20\%$ (i.e. the observed fraction is around $0.2$). In other words, the empirical variance (using MSE) underestimates the true uncertainty. The trend is the same for all datasets. We also report the Expected Calibration Error (ECE) and the Maximum Calibration Error (MCE) \cite{guo2017calibration, naeini2015obtaining} corresponding to MSE Ensemble, Deep Ensemble, and BVM Ensemble on the regression benchmarks datasets. The results agree with those in Figure \ref{figreldiag} and are reported in Table \ref{tablemce}.
\begin{table}[h]
\centering
\caption{The Expected Calibration Error (ECE) and Maximum Calibration Error (MCE) corresponding to MSE Ensemble, Deep Ensemble, and BVM Ensemble on the regression benchmark datasets.\label{tablemce}}
\scalebox{.8}{
\begin{tabular}{l|ccc|ccc}
\toprule
&\multicolumn{3}{ c| }{Expected Calibration Error (ECE)} &\multicolumn{3}{ c }{Maximum Calibration Error (MCE)}\\
\hhline{~~~~~~~}
\textbf{Dataset} & \textbf{MSE Ens.} & \textbf{Deep Ens.} & \textbf{BVM Ens.} & \textbf{MSE Ens.} & \textbf{Deep Ens.} & \textbf{BVM Ens.}\\
\midrule
Boston housing & 0.3010 & 0.0441 & 0.0316 & 0.6059 & 0.1235 & 0.0863\\
Concrete & 0.3888 & 0.0417 & 0.0251 & 0.7447 & 0.1282 & 0.0670\\
Energy & 0.3513 & 0.0209 & 0.0142 & 0.6662 & 0.0506 & 0.0325\\
Kin8nm & 0.3152 & 0.0250 & 0.0626 & 0.6122 & 0.0500 & 0.0976\\
Naval propulsion plant & 0.3556 & 0.0580 & 0.0321 & 0.6680 & 0.1055 & 0.0558\\
Power plant & 0.3934 & 0.0109 & 0.0781 & 0.7704 & 0.0203 & 0.1471\\
Protein & 0.3938 & 0.0222 & 0.0279 & 0.7677 & 0.0408 & 0.0574\\
Wine & 0.3944 & 0.0288 & 0.0445 & 0.7438 & 0.0500 & 0.0875\\
Yacht & 0.4081 & 0.1377 & 0.1412 & 0.7710 & 0.2710 & 0.3055\\
\bottomrule
\end{tabular}
}
\end{table}
\clearpage

\section{Why does BVM outperform Deep Ensembles on outlier samples? (detailed proof)}
\label{AppendixB}
\vspace*{3pt}
Recall from Section \ref{theepseBVMlossfctn} that the $\epsilon$-BVM probability of agreement for a given input feature vector $\mathbf{x}_n$ can be expressed as 
\begin{align}
p\big(A\big|M,D,B(\epsilon),\mathbf{x}_n\big) = \Phi\bigg(\frac{t_n+\epsilon-\mu_n}{\sigma_n}\bigg) - \Phi\bigg(\frac{t_n-\epsilon-\mu_n}{\sigma_n}\bigg),\label{deltacdf}
\end{align}
where $\Phi(\cdot)$ is the cumulative distribution function (cdf) of the standard normal distribution:
\begin{align}
\Phi(x) = \int_{-\infty}^{x}\frac{1}{\sqrt{2\pi}}e^{-x^2/2}dx.
\end{align}
Also recall that the (overall) negative log $\epsilon$-BVM probability of agreement (i.e. the BVM loss function) over the set of all input feature vectors $\mathbf{x} = \{\mathbf{x}_1, \hdots, \mathbf{x}_N\}$ is
\begin{align}
\mathcal{C}_{\text{BVM}}\big(B(\epsilon)\big) &= \frac{1}{N}\sum_{n=1}^N -\log{p\big(A|M,D,B(\epsilon),\mathbf{x}_n\big)}\notag\\[\medskipamount]
&= \frac{1}{N}\sum_{n=1}^N -\log\Bigg[\Phi\bigg(\frac{t_n+\epsilon-\mu_n}{\sigma_n}\bigg) - \Phi\bigg(\frac{t_n-\epsilon-\mu_n}{\sigma_n}\bigg)\Bigg].\label{bvmcdf}
\end{align}
\noindent Note that for a given input feature vector $\mathbf{x}_n$, the minimizer of the BVM loss function satisfies
\begin{align}
&\underset{\mu_n,\sigma_n}{\text{argmin}}\,-\log\Bigg[\Phi\bigg(\frac{t_n+\epsilon-\mu_n}{\sigma_n}\bigg) - \Phi\bigg(\frac{t_n-\epsilon-\mu_n}{\sigma_n}\bigg)\Bigg]\notag\\[\medskipamount]
=\,\, &\underset{\mu_n,\sigma_n}{\text{argmin}}\,-\log\Bigg(\frac{1}{2\epsilon}\bigg[\Phi\bigg(\frac{t_n+\epsilon-\mu_n}{\sigma_n}\bigg) - \Phi\bigg(\frac{t_n-\epsilon-\mu_n}{\sigma_n}\bigg)\bigg]\Bigg)
\end{align}
Taylor expanding around $\epsilon = 0$ leads to
\begin{align}
&-\log\Bigg(\frac{1}{2\epsilon}\bigg[\Phi\bigg(\frac{t_n+\epsilon-\mu_n}{\sigma_n}\bigg) - \Phi\bigg(\frac{t_n-\epsilon-\mu_n}{\sigma_n}\bigg)\bigg]\Bigg) \notag\\[\medskipamount]
&\simeq \frac{1}{2}\log 2\pi\sigma^2_n + \frac{\big(t_n-\mu_n\big)^2}{2\sigma^2_n} - \frac{\epsilon^2}{6}\,\Bigg[\frac{\big(t_n-\mu_n\big)^2}{\sigma^4_n}-\frac{1}{\sigma_n^2}\Bigg] + \mathcal{O}(\epsilon^3)
\end{align}
\begin{proof}
Let for a given input feature vector $\mathbf{x}_n$ the function $g(\epsilon,\mathbf{x}_n)$ be defined by
\begin{align}
g(\epsilon,\mathbf{x}_n) = \frac{1}{2\epsilon}\,p\big(A\big|M,D,B(\epsilon),\mathbf{x}_n\big) = \frac{1}{2\epsilon}\bigg[\Phi\bigg(\frac{t_n+\epsilon-\mu_n}{\sigma_n}\bigg) - \Phi\bigg(\frac{t_n-\epsilon-\mu_n}{\sigma_n}\bigg)\bigg]
\end{align}
Then, we have
\small
\begin{align}
g'(\epsilon,\mathbf{x}_n) &= \frac{1}{4\epsilon^2}\Bigg(2\epsilon\bigg[\frac{1}{\sigma_n}\varphi\bigg(\frac{t_n+\epsilon-\mu_n}{\sigma_n}\bigg) + \frac{1}{\sigma_n}\varphi\bigg(\frac{t_n-\epsilon-\mu_n}{\sigma_n}\bigg)\bigg] - 2\bigg[\Phi\bigg(\frac{t_n+\epsilon-\mu_n}{\sigma_n}\bigg) - \Phi\bigg(\frac{t_n-\epsilon-\mu_n}{\sigma_n}\bigg)\bigg]\,\Bigg)\notag\\[\smallskipamount]
&= \frac{1}{2\epsilon^2}\Bigg(\frac{\epsilon}{\sigma_n}\bigg[\varphi\bigg(\frac{t_n+\epsilon-\mu_n}{\sigma_n}\bigg) + \varphi\bigg(\frac{t_n-\epsilon-\mu_n}{\sigma_n}\bigg)\bigg] - \bigg[\Phi\bigg(\frac{t_n+\epsilon-\mu_n}{\sigma_n}\bigg) - \Phi\bigg(\frac{t_n-\epsilon-\mu_n}{\sigma_n}\bigg)\bigg]\,\Bigg)
\end{align}
\normalsize
and
\small
\begin{align}
g''(\epsilon,\mathbf{x}_n) &= \frac{1}{4\epsilon^4}\Bigg[2\epsilon^2\Bigg(\cancel{\frac{1}{\sigma_n}\bigg[\varphi\bigg(\frac{t_n+\epsilon-\mu_n}{\sigma_n}\bigg) + \varphi\bigg(\frac{t_n-\epsilon-\mu_n}{\sigma_n}\bigg)\bigg]} + \frac{\epsilon}{\sigma_n^2}\bigg[\varphi'\bigg(\frac{t_n+\epsilon-\mu_n}{\sigma_n}\bigg) - \varphi'\bigg(\frac{t_n-\epsilon-\mu_n}{\sigma_n}\bigg)\bigg]\notag\\[\smallskipamount]
&\hspace{44.75pt} - \cancel{\frac{1}{\sigma_n}\bigg[\varphi\bigg(\frac{t_n+\epsilon-\mu_n}{\sigma_n}\bigg) + \varphi\bigg(\frac{t_n-\epsilon-\mu_n}{\sigma_n}\bigg)\bigg]}\,\Bigg)\notag\\[\smallskipamount]
&\hspace{24.5pt} -4\epsilon\,\,\Bigg(\frac{\epsilon}{\sigma_n}\bigg[\varphi\bigg(\frac{t_n+\epsilon-\mu_n}{\sigma_n}\bigg) + \varphi\bigg(\frac{t_n-\epsilon-\mu_n}{\sigma_n}\bigg)\bigg] - \bigg[\Phi\bigg(\frac{t_n+\epsilon-\mu_n}{\sigma_n}\bigg) - \Phi\bigg(\frac{t_n-\epsilon-\mu_n}{\sigma_n}\bigg)\bigg]\,\Bigg)\,\Bigg]\notag\\[\smallskipamount]
&= \frac{1}{2\epsilon^3}\Bigg[\frac{\epsilon^2}{\sigma_n^2}\bigg[\varphi'\bigg(\frac{t_n+\epsilon-\mu_n}{\sigma_n}\bigg) - \varphi'\bigg(\frac{t_n-\epsilon-\mu_n}{\sigma_n}\bigg)\bigg]\notag\\[\smallskipamount]
&\hspace{28.5pt} -2\,\Bigg(\frac{\epsilon}{\sigma_n}\bigg[\varphi\bigg(\frac{t_n+\epsilon-\mu_n}{\sigma_n}\bigg) + \varphi\bigg(\frac{t_n-\epsilon-\mu_n}{\sigma_n}\bigg)\bigg] - \bigg[\Phi\bigg(\frac{t_n+\epsilon-\mu_n}{\sigma_n}\bigg) - \Phi\bigg(\frac{t_n-\epsilon-\mu_n}{\sigma_n}\bigg)\bigg]\,\Bigg)\,\Bigg]
\end{align}
\normalsize
where $\varphi(\cdot)$ is the probability density function (pdf) of the standard normal distribution:
\begin{align}
\varphi(x) = \frac{1}{\sqrt{2\pi}}e^{-x^2/2}
\end{align}
In what follow we will also use $\varphi'(\cdot)$ and $\varphi''(\cdot)$ which are expressed as
\begin{align}
\varphi'(x) = -\frac{x}{\sqrt{2\pi}}e^{-x^2/2}\qquad\text{and}\qquad\varphi''(x) = \frac{x^2-1}{\sqrt{2\pi}}e^{-x^2/2} = (x^2 - 1)\,\varphi(x) 
\end{align}
The Taylor series approximation of $g(\epsilon,\mathbf{x}_n)$ near $\epsilon = 0$ is
\begin{align}
g(\epsilon,\mathbf{x}_n) &\simeq g(0,\mathbf{x}_n) + \frac{g'(0,\mathbf{x}_n)}{1!}(\epsilon - 0) + \frac{g''(0,\mathbf{x}_n)}{2!}(\epsilon - 0)^2 + \mathcal{O}(\epsilon^3)\notag\\[\smallskipamount]
&= \underbrace{g(0,\mathbf{x}_n)}_{\text{Term 1}} + \epsilon\, \underbrace{g'(0,\mathbf{x}_n)}_{\text{Term 2}} + \frac{\epsilon^2}{2}\,\underbrace{g''(0,\mathbf{x}_n)}_{\text{Term 3}} + \mathcal{O}(\epsilon^3)
\end{align}
where $g(0,\mathbf{x}_n)$, $g'(0,\mathbf{x}_n)$, and $g''(0,\mathbf{x}_n)$ can be derived as follows:\vspace{5pt}\\
\underline{Term 1:}
\begin{align}
g(0,\mathbf{x}_n) &= \lim_{\epsilon \to 0} g(\epsilon,\mathbf{x}_n)\notag\\[\smallskipamount]
&= \lim_{\epsilon \to 0} \frac{1}{2\epsilon}\bigg[\Phi\bigg(\frac{t_n+\epsilon-\mu_n}{\sigma_n}\bigg) - \Phi\bigg(\frac{t_n-\epsilon-\mu_n}{\sigma_n}\bigg)\bigg]\notag\\[\smallskipamount]
&= \lim_{\epsilon \to 0} \frac{1}{2}\bigg[\frac{1}{\sigma_n}\varphi\bigg(\frac{t_n+\epsilon-\mu_n}{\sigma_n}\bigg) + \frac{1}{\sigma_n}\varphi\bigg(\frac{t_n-\epsilon-\mu_n}{\sigma_n}\bigg)\bigg]\qquad \text{\big(using L'H\^{o}pital's rule\big)}\notag\\[\smallskipamount]
&= \frac{1}{2\sigma_n}\bigg[\varphi\bigg(\frac{t_n-\mu_n}{\sigma_n}\bigg) + \varphi\bigg(\frac{t_n-\mu_n}{\sigma_n}\bigg)\bigg]\notag\\[\smallskipamount]
&= \frac{1}{\sigma_n}\varphi\bigg(\frac{t_n-\mu_n}{\sigma_n}\bigg)
\end{align}
It follows that $g(0,\mathbf{x}_n)$ is the pdf of the general normal distribution:
\begin{align}
\boxed{g(0,\mathbf{x}_n) = \frac{1}{\sqrt{2\pi\sigma^2_n}}\,\text{exp}\,\Bigg\{-\frac{\big(t_n-\mu_n\big)^2}{2\sigma^2_n}\Bigg\}}
\end{align}
\underline{Term 2:}
\small
\begin{align}
g'(0,\mathbf{x}_n) &= \lim_{\epsilon \to 0} g'(\epsilon,\mathbf{x}_n)\notag\\[\smallskipamount]
&= \lim_{\epsilon \to 0} \frac{1}{2\epsilon^2}\Bigg(\frac{\epsilon}{\sigma_n}\bigg[\varphi\bigg(\frac{t_n+\epsilon-\mu_n}{\sigma_n}\bigg) + \varphi\bigg(\frac{t_n-\epsilon-\mu_n}{\sigma_n}\bigg)\bigg] - \bigg[\Phi\bigg(\frac{t_n+\epsilon-\mu_n}{\sigma_n}\bigg) - \Phi\bigg(\frac{t_n-\epsilon-\mu_n}{\sigma_n}\bigg)\bigg]\,\Bigg)\notag\\[\smallskipamount]
&= \lim_{\epsilon \to 0} \frac{1}{4\epsilon}\Bigg(\cancel{\frac{1}{\sigma_n}\bigg[\varphi\bigg(\frac{t_n+\epsilon-\mu_n}{\sigma_n}\bigg) + \varphi\bigg(\frac{t_n-\epsilon-\mu_n}{\sigma_n}\bigg)\bigg]} + \frac{\epsilon}{\sigma_n^2}\bigg[\varphi'\bigg(\frac{t_n+\epsilon-\mu_n}{\sigma_n}\bigg) - \varphi'\bigg(\frac{t_n-\epsilon-\mu_n}{\sigma_n}\bigg)\bigg]\notag\\[\smallskipamount]
&\hspace{37.75pt} - \cancel{\frac{1}{\sigma_n}\bigg[\varphi\bigg(\frac{t_n+\epsilon-\mu_n}{\sigma_n}\bigg) + \varphi\bigg(\frac{t_n-\epsilon-\mu_n}{\sigma_n}\bigg)\bigg]}\,\Bigg)\qquad \text{\big(using L'H\^{o}pital's rule\big)}\notag\\[\smallskipamount]
&= \lim_{\epsilon \to 0} \frac{\cancel{\epsilon}}{4\cancel{\epsilon}\sigma_n^2}\bigg[\varphi'\bigg(\frac{t_n+\epsilon-\mu_n}{\sigma_n}\bigg) - \varphi'\bigg(\frac{t_n-\epsilon-\mu_n}{\sigma_n}\bigg)\bigg]\notag\\[\smallskipamount]
&= \frac{1}{4\sigma_n^2}\bigg[\varphi'\bigg(\frac{t_n-\mu_n}{\sigma_n}\bigg) - \varphi'\bigg(\frac{t_n-\mu_n}{\sigma_n}\bigg)\bigg] = 0
\end{align}\normalsize
It follows that
\begin{align}
\boxed{g'(0,\mathbf{x}_n) = 0}
\end{align}
\underline{Term 3:}
\small
\begin{align}
g''(0,\mathbf{x}_n) &= \lim_{\epsilon \to 0} g''(\epsilon,\mathbf{x}_n)\notag\\[\smallskipamount]
&= \lim_{\epsilon \to 0} \frac{1}{2\epsilon^3}\Bigg[\hspace{0pt}\frac{\epsilon^2}{\sigma_n^2}\bigg[\varphi'\bigg(\frac{t_n+\epsilon-\mu_n}{\sigma_n}\bigg) - \varphi'\bigg(\frac{t_n-\epsilon-\mu_n}{\sigma_n}\bigg)\bigg]\notag\\[\smallskipamount]
&\hspace{45pt} -2\,\Bigg(\frac{\epsilon}{\sigma_n}\bigg[\varphi\bigg(\frac{t_n+\epsilon-\mu_n}{\sigma_n}\bigg) + \varphi\bigg(\frac{t_n-\epsilon-\mu_n}{\sigma_n}\bigg)\bigg] - \bigg[\Phi\bigg(\frac{t_n+\epsilon-\mu_n}{\sigma_n}\bigg) - \Phi\bigg(\frac{t_n-\epsilon-\mu_n}{\sigma_n}\bigg)\bigg]\,\Bigg)\,\Bigg]\notag\\[\smallskipamount]
&= \lim_{\epsilon \to 0} \frac{1}{6\epsilon^2}\Bigg[\cancel{\frac{2\epsilon}{\sigma_n^2}\bigg[\varphi'\bigg(\frac{t_n+\epsilon-\mu_n}{\sigma_n}\bigg) - \varphi'\bigg(\frac{t_n-\epsilon-\mu_n}{\sigma_n}\bigg)\bigg]} + \frac{\epsilon^2}{\sigma_n^3}\bigg[\varphi''\bigg(\frac{t_n+\epsilon-\mu_n}{\sigma_n}\bigg) + \varphi''\bigg(\frac{t_n-\epsilon-\mu_n}{\sigma_n}\bigg)\bigg]\notag\\[\smallskipamount]
&\hspace{45pt} -2\,\Bigg(\cancel{\frac{1}{\sigma_n}\bigg[\varphi\bigg(\frac{t_n+\epsilon-\mu_n}{\sigma_n}\bigg) + \varphi\bigg(\frac{t_n-\epsilon-\mu_n}{\sigma_n}\bigg)\bigg]} + \cancel{\frac{\epsilon}{\sigma_n^2}\bigg[\varphi'\bigg(\frac{t_n+\epsilon-\mu_n}{\sigma_n}\bigg) - \varphi'\bigg(\frac{t_n-\epsilon-\mu_n}{\sigma_n}\bigg)\bigg]} \notag\\[\smallskipamount]
&\hspace{59.8pt}- \cancel{\frac{1}{\sigma_n}\bigg[\varphi\bigg(\frac{t_n+\epsilon-\mu_n}{\sigma_n}\bigg) + \varphi\bigg(\frac{t_n-\epsilon-\mu_n}{\sigma_n}\bigg)\bigg]}\,\Bigg)\,\Bigg]\qquad \text{\big(using L'H\^{o}pital's rule\big)}\notag\\[\smallskipamount]
&= \frac{\cancel{\epsilon^2}}{6\cancel{\epsilon^2}\sigma_n^3}\bigg[\varphi''\bigg(\frac{t_n-\mu_n}{\sigma_n}\bigg) + \varphi''\bigg(\frac{t_n-\mu_n}{\sigma_n}\bigg)\bigg]\notag\\[\smallskipamount]
&= \frac{1}{3\sigma_n^3}\varphi''\bigg(\frac{t_n-\mu_n}{\sigma_n}\bigg)\notag\\[\smallskipamount]
&= \frac{1}{3\sigma_n^3}\Bigg[\frac{\big(t_n-\mu_n\big)^2}{\sigma^2_n}-1\Bigg]\varphi\bigg(\frac{t_n-\mu_n}{\sigma_n}\bigg)\qquad\text{\big(using $\varphi''(x) = (x^2 - 1)\,\varphi(x)$\big)}
\end{align}\normalsize
It follows that
\begin{align}
\boxed{g''(0,\mathbf{x}_n) = \frac{1}{3\sigma_n^2}\Bigg[\frac{\big(t_n-\mu_n\big)^2}{\sigma^2_n}-1\Bigg]\frac{1}{\sqrt{2\pi\sigma^2_n}}\,\text{exp}\,\Bigg\{-\frac{\big(t_n-\mu_n\big)^2}{2\sigma^2_n}\Bigg\}}
\end{align}\noindent\vspace{5pt}\\
Hence, the Taylor series approximation of $g(\epsilon,\mathbf{x}_n)$ around $\epsilon = 0$ is
\small
\begin{align}
g(\epsilon,\mathbf{x}_n) &\simeq g(0,\mathbf{x}_n) + \epsilon\, g'(0,\mathbf{x}_n) + \frac{\epsilon^2}{2}\,g''(0,\mathbf{x}_n) + \mathcal{O}(\epsilon^3)\notag\\[\medskipamount]
&= \frac{1}{\sqrt{2\pi\sigma^2_n}}\,\text{exp}\,\Bigg\{-\frac{\big(t_n-\mu_n\big)^2}{2\sigma^2_n}\Bigg\} + \frac{\epsilon^2}{2}\,\frac{1}{3\sigma_n^2}\Bigg[\frac{\big(t_n-\mu_n\big)^2}{\sigma^2_n}-1\Bigg]\frac{1}{\sqrt{2\pi\sigma^2_n}}\,\text{exp}\,\Bigg\{-\frac{\big(t_n-\mu_n\big)^2}{2\sigma^2_n}\Bigg\} + \mathcal{O}(\epsilon^3)\notag\\[\medskipamount]
&= \frac{1}{\sqrt{2\pi\sigma^2_n}}\,\text{exp}\,\Bigg\{-\frac{\big(t_n-\mu_n\big)^2}{2\sigma^2_n}\Bigg\}\Bigg(1 + \frac{\epsilon^2}{2}\,\frac{1}{3\sigma_n^2}\Bigg[\frac{\big(t_n-\mu_n\big)^2}{\sigma^2_n}-1\Bigg] + \mathcal{O}(\epsilon^3)\Bigg)
\end{align}
\normalsize
Taking its negative log gives
\small
\begin{align}
-\log g(\epsilon,\mathbf{x}_n) &\simeq \frac{1}{2}\log 2\pi\sigma^2_n + \frac{\big(t_n-\mu_n\big)^2}{2\sigma^2_n} -\log\Bigg(1 + \frac{\epsilon^2}{2}\,\frac{1}{3\sigma_n^2}\Bigg[\frac{\big(t_n-\mu_n\big)^2}{\sigma^2_n}-1\Bigg] + \mathcal{O}(\epsilon^3)\Bigg)\notag\\[\smallskipamount]
&\simeq \frac{1}{2}\log 2\pi\sigma^2_n + \frac{\big(t_n-\mu_n\big)^2}{2\sigma^2_n} - \frac{\epsilon^2}{2}\,\frac{1}{3\sigma_n^2}\Bigg[\frac{\big(t_n-\mu_n\big)^2}{\sigma^2_n}-1\Bigg] + \mathcal{O}(\epsilon^3)\qquad\text{\big($\log(1+x)\simeq x$ for $x$ near $0$\big)}\notag\\[\smallskipamount]
&= \frac{1}{2}\log 2\pi\sigma^2_n + \frac{\big(t_n-\mu_n\big)^2}{2\sigma^2_n} - \frac{\epsilon^2}{6}\,\Bigg[\frac{\big(t_n-\mu_n\big)^2}{\sigma^4_n}-\frac{1}{\sigma_n^2}\Bigg] + \mathcal{O}(\epsilon^3)
\end{align}
\normalsize
\end{proof}
\noindent Thus, the minimizer of the BVM loss over the set of all input feature vectors can be approximated as
\begin{align}
\underset{\mu_n,\sigma_n}{\text{argmin}}\,\underbrace{\frac{1}{N}\sum_{n=1}^N\Bigg(\frac{1}{2}\log 2\pi\sigma^2_n + \frac{\big(t_n-\mu_n\big)^2}{2\sigma^2_n}}_{\mathcal{C}_{\text{NLL}}} - \frac{\epsilon^2}{6}\,\Bigg[\frac{\big(t_n-\mu_n\big)^2}{\sigma^4_n}-\frac{1}{\sigma_n^2}\Bigg]\Bigg)
\end{align}
We can clearly see that for $\epsilon = 0$, the minimizer of the BVM loss is indeed the minimizer of the NLL loss in Equation (\ref{nll}). For a nonzero $\epsilon$, $\sigma_n$ will increase linearly with $\epsilon$ (see proof below) leading to a larger variance, and hence a wider distribution (or prediction interval). Thus, the outlier samples near the tails will be more probable resulting in lower NLL values compared to Deep Ensembles (keeping in mind that the samples near the mean will be less probable resulting in higher NLL values compared to Deep Ensembles).
\begin{proof}
Let for a given input feature vector $\mathbf{x}_n$ the function $f(\epsilon,\mathbf{x}_n)$ be defined by
\begin{align}
f(\epsilon,\mathbf{x}_n) = \frac{1}{2}\log 2\pi\sigma^2_n + \frac{\big(t_n-\mu_n\big)^2}{2\sigma^2_n} - \frac{\epsilon^2}{6}\,\Bigg[\frac{\big(t_n-\mu_n\big)^2}{\sigma^4_n}-\frac{1}{\sigma_n^2}\Bigg]
\end{align}
For a fixed $\epsilon$, the minimizers $\mu_n$ and $\sigma_n$ can be found by computing the gradients $\nabla_{\mu_n} f$ and $\nabla_{\sigma_n} f$ and setting them to zero:
\begin{align}
\nabla_{\mu_n} f &= - \frac{t_n - \mu_n}{\sigma_n^2} + \frac{\epsilon^2}{3}\frac{t_n-\mu_n}{\sigma_n^4} = \frac{t_n - \mu_n}{\sigma_n^2}\bigg(-1 + \frac{\epsilon^2}{3\sigma_n^2}\bigg)\\[\smallskipamount]
\nabla_{\sigma_n} f &= \frac{1}{\sigma_n} - \frac{(t_n - \mu_n)^2}{\sigma_n^3} - \frac{\epsilon^2}{6}\,\Bigg[-4\,\frac{\big(t_n-\mu_n\big)^2}{\sigma^5_n}+\frac{2}{\sigma_n^3}\Bigg]
\end{align}
Note that
\begin{align}
\nabla_{\mu_n} f = 0\qquad \text{for}\qquad \underbrace{\mu_n = t_n}_{\text{Case 1}} \qquad\text{or}\qquad \underbrace{\sigma_n =  \epsilon/\sqrt{3}}_{\text{Case 2}}
\end{align}
\underline{Case 1:} \,\,$\mu_n = t_n$\vspace{5pt}\\
In this case, we set $\nabla_{\sigma_n} f$ to zero and we get
\begin{align}
\nabla_{\sigma_n} f &= \frac{1}{\sigma_n} - \cancelto{0}{\frac{(t_n - \mu_n)^2}{\sigma_n^3}} - \frac{\epsilon^2}{6}\,\Bigg[-4\,\cancelto{0}{\frac{\big(t_n-\mu_n\big)^2}{\sigma^5_n}}+\frac{2}{\sigma_n^3}\Bigg] = \frac{1}{\sigma_n} - \frac{\epsilon^2}{3\sigma_n^3} = 0 \quad \Longrightarrow \quad \sigma_n = \epsilon/\sqrt{3}
\end{align}
\underline{Case 2:} \,\,$\sigma_n =  \epsilon/\sqrt{3}$\vspace{5pt}\\
In this case, we set $\nabla_{\sigma_n} f$ to zero and we get
\begin{align}
\nabla_{\sigma_n} f &= \frac{1}{\sigma_n} - \frac{(t_n - \mu_n)^2}{\sigma_n^3} - \frac{\epsilon^2}{6}\,\Bigg[-4\,\frac{\big(t_n-\mu_n\big)^2}{\sigma^5_n}+\frac{2}{\sigma_n^3}\Bigg]\notag\\[\smallskipamount]
&= \frac{1}{\sigma_n} - \frac{(t_n - \mu_n)^2}{\sigma_n^3}\bigg[1-2\cancelto{1}{\frac{\epsilon^2}{3\sigma_n^2}}\,\,\bigg] - \frac{1}{\sigma_n}\cancelto{1}{\frac{\epsilon^2}{3\sigma_n^2}}\notag\\[\smallskipamount]
&= \cancel{\frac{1}{\sigma_n}} - \frac{(t_n - \mu_n)^2}{\sigma_n^3}(-1) - \cancel{\frac{1}{\sigma_n}}\notag\\[\smallskipamount]
&= \frac{(t_n - \mu_n)^2}{\sigma_n^3} = 0 \quad \Longrightarrow \quad \mu_n = t_n
\end{align}
Thus, in both cases, we have $\mu_n = t_n$ and $\sigma_n = \epsilon/\sqrt{3}$. \\
It follows that $\sigma_n$ increases linearly with $\epsilon$. The larger $\epsilon$, the larger the variance and hence the more probable the outlier samples (and the less probable the nonoutlier samples).\\ 
\end{proof}

\end{document}